\documentclass[runningheads]{llncs}
\usepackage{graphicx}
\usepackage{amsmath,amsfonts,bm}

\def\eqref#1{equation~\ref{#1}}

\def\1{\bm{1}}

\def\vx{{\bm{x}}}

\def\vz{{\bm{z}}}

\DeclareMathAlphabet{\mathsfit}{\encodingdefault}{\sfdefault}{m}{sl}
\SetMathAlphabet{\mathsfit}{bold}{\encodingdefault}{\sfdefault}{bx}{n}

\newcommand{\R}{\mathbb{R}}

\usepackage{multirow}
\usepackage{nicefrac}
\usepackage{caption}
\usepackage{subfigure}
\usepackage{array}
\usepackage{wrapfig}
\usepackage{hyperref}
\usepackage{url}
\usepackage{microtype}
\usepackage{graphicx}
\usepackage{subfigure}
\usepackage{booktabs} 

\usepackage[misc]{ifsym} 

\begin{document}
\title{
Invertible Manifold Learning for Dimension Reduction
}
%
%

\author{
Siyuan Li\inst{1,2}\orcidID{0000-0001-6806-2468} \and
Haitao Lin\inst{1,2} \and
Zelin Zang\inst{1,2} \and\\
Lirong Wu\inst{1,2} \and
Jun Xia\inst{1,2} \and
Stan Z. Li\inst{1,2}\orcidID{0000-0002-2961-8096}\thanks{Corresponding author.}
}
\authorrunning{S. Li et al.}
%
\institute{
AI Lab, School of Engineering, Westlake University, Hangzhou, Zhejiang, China \and
Institute of Advanced Technology, Westlake Institute for Advanced Study,\\
Hangzhou, Zhejiang, China
\email{\{lisiyuan,linhaitao,zangzelin,wulirong,xiajun,Stan.ZQ.Li\}@westlake.edu.cn}}

\maketitle              

\begin{abstract}
Dimension reduction (DR) aims to learn low-dimensional representations of high-dimensional data with the preservation of essential information. In the context of manifold learning, we define that the representation after information-lossless DR preserves the topological and geometric properties of data manifolds formally, and propose a novel two-stage DR method, called invertible manifold learning (\textit{inv-ML}) to bridge the gap between theoretical information-lossless and practical DR. The first stage includes a homeomorphic \textit{sparse coordinate transformation} to learn low-dimensional representations without destroying topology and a \textit{local isometry} constraint to preserve local geometry. In the second stage, a \textit{linear compression} is implemented for the trade-off between the target dimension and the incurred information loss in excessive DR scenarios. 
Experiments are conducted on seven datasets with a neural network implementation of \textit{inv-ML}, called \textit{i-ML-Enc}. Empirically, \textit{i-ML-Enc} achieves invertible DR in comparison with typical existing methods as well as reveals the characteristics of the learned manifolds. Through latent space interpolation on real-world datasets, we find that the reliability of tangent space approximated by the local neighborhood is the key to the success of manifold-based DR algorithms.

\end{abstract}

%
%
%
\section{Introduction}

In real-world scenarios, it is widely believed that the loss of data information is inevitable after dimension reduction (DR), though the goal of DR is to preserve as much data information as possible in the low-dimensional space. Most methods try to preserve some essential information of data after DR, e.g., geometric structure within the data, which is usually achieved by preserving the distance in high and low-dimensional space. In the case of linear DR, compressed sensing \cite{2006-CS} breaks this common sense with practical sparse conditions of the given data. The lower bound of target dimension and the information loss for linear DR are provided by Johnson–Lindenstrauss Theorem \cite{1984-JL-lamma} with the pairwise distance. In the case of nonlinear dimension reduction (NLDR), however, it has not been thoroughly discussed, i.e., what structures within data are necessary to preserve, how to maintain these structures after NLDR, and how much information can be preserved under different cases? From the perspective of manifold learning, a popular \textit{manifold assumption} is widely adopted that the given data has relatively low-dimensional intrinsic structures. Classical manifold-based DR methods \cite{2000-science-LLE} \cite{2007-MLLE} work well on synthetic manifold datasets, but usually fail to yield good results in the many practical cases. Therefore, there is still a gap between theoretical and real-world applications of manifold-based DR.

Here, we give a detailed discussion of these problems in the context of manifold learning and define that the representation after information-lossless DR should preserve the topology and geometry of input data. On the one hand, the representation should demonstrate some geometric properties after DR, or it will be meaningless. For example, if the distance between point $A$ and $B$ is larger than that between $A$ and $C$ on the data manifold, the low-dimension representation should preserve the order to revealing the simalarity of data points. On the other hand, the topological properties can be preserved if the DR transformation is a continuous bijective mapping, i.e., homeomorphism, leading to the information-lossless mapping.

To achieve the information-lossless DR, we propose an invertible NLDR process, called \textit{inv-ML}, combining \textit{sparse coordinate transformation} and \textit{local isometry} constraint which preserve the property of topology and geometry respectively. In terms of the target dimension and information loss, we discuss different cases of NLDR in manifold learning. We instantiate \textit{inv-ML} as a neural network called \textit{i-ML-Enc} via a cascade of equidimensional layers and a linear transform layer. The proposed loss terms and network structures are explainable. Sufficient experiments are conducted to validate invertible NLDR abilities of \textit{i-ML-Enc} and analyze learned representations to reveal inherent difficulties of classical manifold learning empirically.

We summarize our main contributions as follows:
\setlength{\itemsep}{0pt}
\setlength{\parsep}{0pt}
\setlength{\parskip}{0pt}
\begin{itemize}
\setlength{\itemsep}{0pt}
\setlength{\parsep}{0pt}
\setlength{\parskip}{0pt}
    \item Introduce an invertible NLDR process \textit{inv-ML} to fill the gap between theoretical information-lossless and real-world applications of NLDR.
    \item Verify the proposed \textit{inv-ML} in different cases by designing an invertible neural network \textit{i-ML-Enc} which produces explainable NLDR results and achieves state-of-the-art performance on benchmark datasets.
    \item Reveals characteristics of the learned low-dimensional representation by latent space interpolation.
\end{itemize}

\section{Related Work}
\label{CH_2}
\paragraph{Manifold learning.}
Most classical DR or NLDR methods aim to preserve the geometric properties of manifolds. The Isomap \cite{2000-science-Isomap} based methods aim to preserve the global metric between every pair of sample points. For example, \cite{2016-NIPS-RR} can be regarded as such methods based on the push-forward Riemannian metric. For the other aspect, LLE \cite{2000-science-LLE} based methods try to preserve local geometry after DR, whose derivatives like LTSA \cite{2004-SIAM-LTSA}, MLLE \cite{2007-MLLE}, etc. have been widely used but usually fail in the high-dimensional case. Recently, based on local properties of manifolds, MLDL \cite{2020-MLDL} was proposed as a robust NLDR method implemented by a neural network. However, those methods ignore the retention of topology. In contrast, we take the preservation of both geometry and topology into consideration, trying to maintain these properties of manifolds even in cases of excessive dimension reduction when the target dimension $s'$ is smaller than $s$.

\paragraph{Invertible model.}
From AutoEncoder (AE) \cite{2006-science-AE}, the fundamental neural network based model, having achieved DR and cut information loss by minimizing the reconstruction loss, some AE based generative models like VAE \cite{2014-ICLR-VAE} and manifold-based NLDR models like TopoAE \cite{2020-ICML-TopoAE} and GRAE \cite{2020-GRAE} have emerged. These methods cannot avoid information loss after NLDR, and thus, some invertible models consist of a series of equidimensional layers have been proposed, some of which aim to generate samples by density estimation through layers \cite{2015-ICLR-NICE} \cite{2017-ICLR-RealNVP}  \cite{2019-ICML-iResNet}, and the other of which are established for other targets, e.g., validating the mutual information bottleneck \cite{2018-ICLR-iRevNet}. Different from the methods mentioned above, our proposed \textit{i-ML-Enc} is a neural network based encoder, with NLDR as well as maintaining structures of raw data points based on manifold assumption via a series of equidimensional layers.

\section{Proposed Method}

Firstly, we state the information-lossless DR problem in Section \ref{CH_3.1}. Then, the proposed invertible NLDR process \textit{inv-ML} is specifically discussed in Section \ref{CH_3.2} and Section \ref{CH_3.3}. Finally, we instantiate the proposed \textit{inv-ML} as \textit{i-ML-Enc} in Section \ref{CH_3.4}.

\subsection{Problem Statement}
\label{CH_3.1}
To start, we first make out a theoretical definition of information-lossless DR of a data manifold. The structures of the manifold from which data points are sampled from include topology and geometry, if the transformed manifold preserves these two structures after a dimension reduction process, this DR process is defined as information-lossless.

\paragraph{Topology preservation.}
The topological property is what is invariant under a homeomorphism, and thus what we want to achieve is to construct a homeomorphism for dimension reduction, removing the redundant dimensions while preserving invariant topology. To be more specific, $f:\mathcal{M}_0^d \rightarrow \mathbb{R}^m$ is a smooth mapping of a differential manifold into another, and if $f$ is a homeomorphism of $\mathcal{M}_0^d$ into $ \mathcal{M}_1^d=f(\mathcal{M}_0^d) \subset \mathbb{R}^m$, we call $f$ is an embedding of $\mathcal{M}_0^d$ into $\mathbb{R}^m$. Assume that the data set $\mathcal{X} = \{\vx_j| 1\leq j \leq n\}$ sampled from the compact manifold $\mathcal{M}_1^d \subset \mathbb{R}^m$ which we call the data manifold and is homeomorphic to $\mathcal{M}_0^d$. For the sample points we get are represented in the coordinate after inclusion mapping $i_1$, we can only regard them as points from Euclidean space $\mathbb{R}^m$ without any prior knowledge, and learn to approximate the data manifold in the latent space $Z$. According to the Whitney Embedding Theorem \cite{2016-Whitney-Nash}, $\mathcal{M}_0^d$ can be embedded smoothly into $\mathbb{R}^{2d}$ by a homeomorphism $g$. Rather than to find the $f^{-1}: \mathcal{M}_1^d \rightarrow \mathcal{M}_0^d$, our goal is to seek a smooth map $h : \mathcal{M}_1^d \rightarrow \mathbb{R}^{s} \subset \mathbb{R}^{2d}$, where $h = g \circ f^{-1}$ is a homeomorphism of $\mathcal{M}_1^d$ into $\mathcal{M}_2^d = h(\mathcal{M}_1^d)$ and $d \leq s \leq 2d \ll m$, and thus the $dim(h(\mathcal{X})) = s$, which achieves the DR while preserving the topology. Owing to the homeomorphism $h$ we seek as a DR mapping, the data manifold $\mathcal{M}_1^d$ is reconstructible via $\mathcal{M}_1^d = h^{-1}\circ h(\mathcal{M}_1^d)$, by which we mean $h$ a topology preserving DR as well as information-lossless DR.

\begin{figure}[htb]
  \centering
  \includegraphics[width=3.25in]{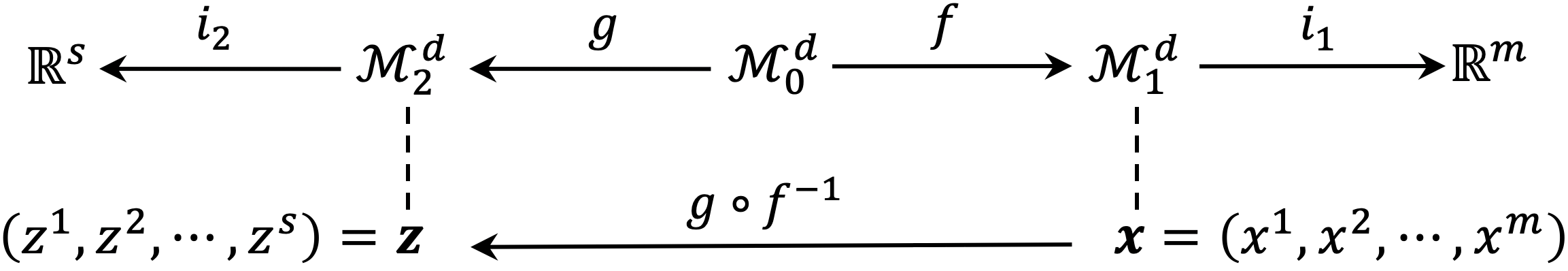}
  \caption{Illustration of the process of NLDR. The dash line links $\mathcal{M}_1^{d}$ and $\bm{x}$ means $\bm{x}$ is sampled from $\mathcal{M}_1^{d}$, and it is represented in the Euclidean space $\mathbb{R}^m$ after an inclusion mapping $i_i$. We aim to approximate $\mathcal{M}_1^{d}$ from the observed sample $\bm{x}$. For the topology preserving dimension reduction methods, it aims to find a homeomorphism $g\circ f^{-1}$ to map $\vx$ into $\vz$ which is embedded in $\mathbb{R}^s$.}
  \label{figure:ch1_topodr}
\end{figure}

\paragraph{Geometry preservation.}
While the topology of the data manifold $\mathcal{M}_1^d$ can be preserved by the homeomorphism $h$ discussed above, it may distort the geometry. To preserve the local geometry of the data manifold, we choose pair-wise distance as the key geometric property, i.e. the DR mapping should be isometric on the tangent space $\mathcal{T}_p\mathcal{M}_1^d$ for every $p\in \mathcal{M}_1^d$, indicating that $d_{\mathcal{M}_1^d}(u,v) = d_{\mathcal{M}_2^d}(h(u), h(v))$, $\forall u,v\in\mathcal{T}_p\mathcal{M}_1^d$. By Nash's Embedding Theorem \cite{1956-Nash}, any smooth manifold of class $C^{k}$ with $k\geq 3$ and dimension $d$ can be embedded isometrically in the Euclidean space $\mathbb{R}^s$ with $s$ polynomial in $d$.

\paragraph{Noise perturbation.}
In the real-world scenarios, sample points are not lied on the ideal manifold strictly due to the limitation of sampling, e.g., non-uniform sampling noises. When the DR method is very robust to the noise, it is reasonable to ignore the effects of the noise and learn the representation $Z$ from the given data. Therefore, the intrinsic dimension of $\mathcal{X}$ is approximate to $d$, resulting in the lowest isometric embedding dimension is larger than $s$.

\subsection{Methods for Structure Preservation}
\label{CH_3.2}
\paragraph{Canonical embedding for homeomorphism.}
To seek the smooth homeomorphism $h$, we turn to the theorem of local canonical form of immersion \cite{2013-Mei}. Let $f:\mathcal{M} \rightarrow \mathcal{N}$ an immersion, and for any $ p \in \mathcal{M}$, there exist local coordinate systems $(U,\phi)$ around $p$ and $(V,\psi)$ around $f(p)$ such that $\psi \circ f \circ \phi^{-1} : \phi(U) \rightarrow \psi(V)$ is a canonical embedding, which reads
\begin{align}
   \psi \circ f \circ \phi^{-1}(x^1, x^2, \ldots , x^d) = (x^1, x^2, \ldots , x^d, 0, 0, \ldots, 0).
   \label{Eq.CH_3.1-1}
\end{align}
In our case, let $\mathcal{M} = \mathcal{M}_2^d$, and $\mathcal{N} = \mathcal{M}_1^d$, any point $\vz = (z^1, z^2,\ldots, z^s) \in \mathcal{M}_1^d \subset \mathbb{R}^{s}$ can be mapped to a point in $\mathbb{R}^m$ by the canonical embedding
\begin{align}
    \psi \circ h^{-1} (z^1, z^2,\ldots, z^s) = (z^1, z^2,\ldots, z^s, 0, 0, \ldots, 0).
    \label{Eq.CH_3.1-2}
\end{align}
For the point $\vz$ is regarded as a point in $\mathbb{R}^s$, $\phi = \mathbb{I}$ is an identity mapping, and for $h=g\circ f^{-1}$ is a homeomorphism, $h^{-1}$ is continuous. The Eq. (\ref{Eq.CH_3.1-2}) can be written as
\begin{align}
    (z^1, z^2,\ldots, z^s) &= h\circ\psi^{-1}(z^1, z^2,\ldots, z^s, 0, 0, \ldots, 0)\notag \\
                           &= h(x^1, x^2, \ldots, x^m).
                           \label{Eq.CH_3.1-3}
\end{align}
Therefore, to reduce $dim(\mathcal{X}) = m$ to $s$, we can decompose $h$ into $\psi$ and $h\circ \psi^{-1}$, by firstly finding a homeomorphic coordinate transformation $\psi$ to map $\vx=(x^1,x^2,\ldots,x^m)$ into $\psi(\vx) = (z^1, z^2, \ldots, z^s, 0, 0, \ldots, 0) $, which is called a \textit{sparse coordinate transformation}, and $h\circ\psi^{-1}$ can be easily obtained by Eq. (\ref{Eq.CH_3.1-2}). We denote $h\circ\psi^{-1}$ by $h_0$ and call it a \textit{sparse compression}.  The theorem holds for any manifold, while in our case, we aims to find the mapping of $\mathcal{X} \subset \mathbb{R}^m$ into $\mathbb{R}^s$, so the local coordinate systems can be extended to the whole space of $\mathbb{R}^m$. 

\paragraph{Local isometry constraint.} The prior local isometry constraint is applied under the manifold assumption, which aims to preserve distances (or some other metrics) locally so that $d_{\mathcal{M}_1^d}(u,v) = d_{\mathcal{M}_2^d}(h(u), h(v))$, $\forall u,v\in\mathcal{T}_p\mathcal{M}_1^d$.

\subsection{Linear Compression}
\label{CH_3.3}
With the former discussed method, manifold-based NLDR can be achieved with topology and geometry preserved, i.e. $s$-sparse representation in $\R^m$. However, the target dimension $s'$ may be even less than $s$, further compression can be performed through the \textit{linear compression} $h'_0: \R^{m}\rightarrow \R^{s'}$ instead of \textit{sparse compression}, where $h'_0(\vz) = W_{m\times s'}\vz$, with minor information loss. In general, the \textit{sparse compression} is a particular case of \textit{linear compression} with $h_0(\vz) = h'_0(\vz) = \Lambda \vz$, where $\Lambda = (\delta_{i,j})_{m\times s}$ and $\delta_{i,j}$ is the Kronecker delta. We discusses the information loss caused by a linear compression under different target dimensions $s'$ as following cases.

\paragraph{Ideal case.}
In the case of $d\le s\le s'$, based on compressed sensing, we can reconstruct the raw input data after the NLDR process without loss of any information by solving the sparse optimization problem mentioned in Section \ref{CH_2} when the transformation matrix $W_{m\times s'}$ has the full rank of the column. In the case of $d\le s'<s$, it is inevitable to drop the topological properties because the two spaces before and after NLDR are not homeomorphic. It is reduced to local geometry preservation by LIS constraint. However, in the case of $s'\le d<s$, both topological and geometric information is lost to varying degrees. Therefore, we can only try to retain as much geometric structure as possible.
\begin{wrapfigure}{r}{0.42\linewidth}
    \centering
    \includegraphics[width=0.34\textwidth]{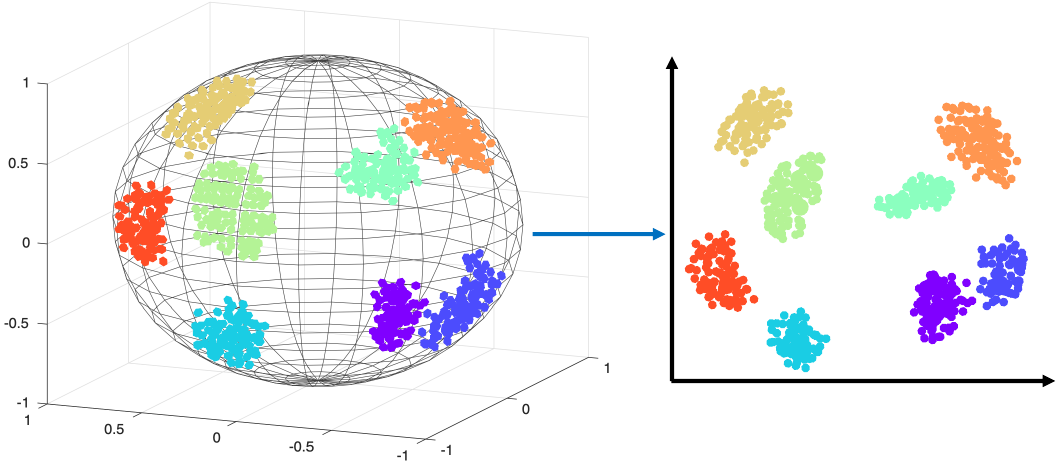}
    \caption{Assuming data points are non-uniform sampled from a high-dimensional hypersphere, it is no need to maintain the global topology for the sparsity and clustering effect.}
\end{wrapfigure}

\paragraph{Practical case.}
In real-world scenarios, the target dimension $s'$ is usually lower than $s$, even lower than $d$. Meanwhile, the data sampling rate is quite low, and the clustering effect is extremely significant, indicating that it is possible to approximate $\mathcal{M}_{1}$ by low-dimensional hyperplane in the Euclidean space. In the case of $s'< s$, we can retain the prior Euclidean topological structure as additional topological information of raw data points. It is reduced to replace the global topology with some relative structures between each cluster.

\begin{figure}[!htb]
  \centering
  \includegraphics[width=3.60in]{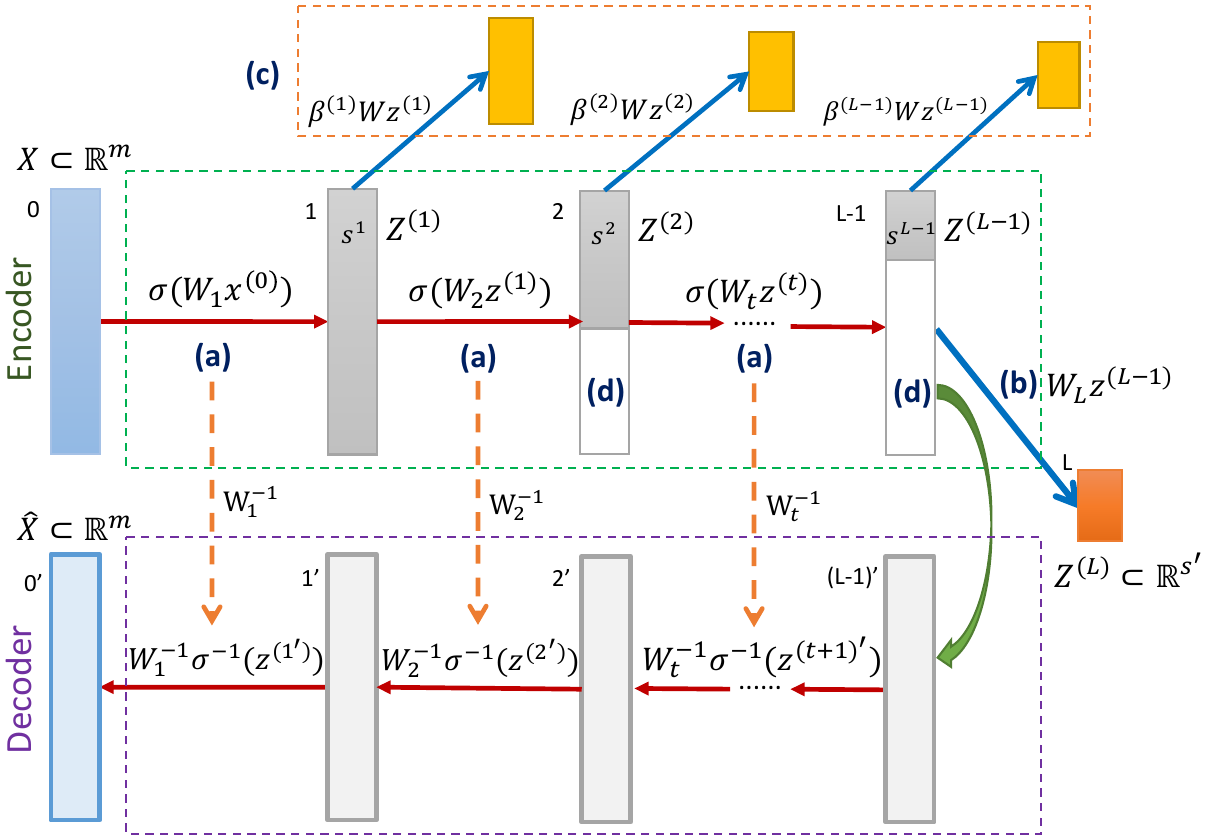}
  \caption{Architecture of the proposed neural network implementation \textit{i-ML-Enc}. The first $L-1$ layers equidimensional mapping in the green dash box are the first stage that achieves $s$-sparse, and they have an inverse process in the purple dash box. (a) denotes a layer of nonlinear homeomorphism transformation (red arrow). (b) linearly transforms (blue arrow) $s$-sparse representation in $\R^m$ into $\R^{s'}$ as the second stage. (c) represents a \textit{extra head} by linear transformations, which will be removed after training. (d) indicates the padding zeros of the $l$-th layer to force $d^{(l)}$-sparse.}
  \label{figure:ch3_network}
\end{figure}

\subsection{Network Implementation}
\label{CH_3.4}
Based on Section \ref{CH_3.2} and Section \ref{CH_3.3}, we propose a neural network \textit{i-ML-Enc} which achieves two-stage NLDR preserving both topology and geometry, as shown in Fig. \ref{figure:ch3_network}. In this section, we will introduce the function of proposed network structures and loss terms respectively, including the orthogonal loss, padding loss, and \textit{extra heads} for the first stage, the LIS loss and push-away loss for the second stage.

\paragraph{Cascade of homeomorphisms.}
Since the \textit{sparse coordinate transformation} $\psi$ (and its inverse) can be highly nonlinear and complex, we decompose it into a cascade of $L-1$ isometric homeomorphisms $\mathrm{\psi}=\ \psi^{(L-1)}\circ\ldots\circ\psi^{(2)}\circ {\psi}^{(1)}$, which can be achieved by  $L-1$ equidimensional network layers. For each ${\psi}^{(l)}$, it is a \textit{sparse coordinate transformation}, where $\psi^l(z^{1,(l)},z^{2,(l)},\ldots,z^{s_l,(l)}, 0,\ldots,0) =(z^{1,(l+1)},z^{2,(l+1)},\ldots,z^{s_{l+1},(l+1)}, 0,\ldots,0)$ with $s_{l+1}<s_{l}$ and $s_{L-1} = s$. The layer-wise transformation $Z^{(l+1)}=\psi^{(l)}( Z^{(l)})$ and its inverse can be written as
\begin{align}
	Z^{(l+1)} = \sigma(W_{l}X^{(l)}),\ Z^{(l)^{'}}=W^{-1}_{l}(\sigma^{-1}(Z^{(l+1)^{'}})), \label{Eq.CH_3.3-1}
\end{align}
in which $W_{l}$ is the $l$-th weight matrix of the neural network to be learned, and $\sigma(.)$ is a nonlinear activation. The bias term is removed here to facilitate its simple inverse structure.

\paragraph{Orthogonal loss.}
Each layer-wise transformation is thought to be a homeomorphism between $Z^{(l)}$ and $Z^{(l+1)}$ in the first $L-1$ layers, and we want it to be a nearly isometric as
\begin{align}
	(1-\epsilon)\Vert \vx_1-\vx_2\Vert \le \Vert W(\vx_1-\vx_2)\Vert \le (1+\epsilon)\Vert \vx_1-\vx_2\Vert,
	\label{Eq.RIP}
\end{align}
where $\epsilon \in(0,1)$ is a rather small constant and $W$ is a linear measurement of signal $\vx_1$ and $\vx_2$. Because the activation function $\sigma(.)$ is monotonous, we can rewrite Eq.(\ref{Eq.RIP}) as
\begin{align}
	L_{orth} = \sum_{l=1}^{L-1} \alpha^{(l)} \rho(W_{l}^TW_{l} - I),
	\label{Eq.CH_3.3-orth}
\end{align}
where $\{\alpha^{(l)}\}$ are the loss weights. Notice that $\rho(W)=\sup_{\vz\in\R^m,\vz\neq\textbf{0}}\frac{\vert W\vz\vert}{\vert \vz\vert}$ is the spectral norm of $W$, and the loss term can be written as $\rho(W_{l}^TW_{l}-I) = \sup_{\vz\in\R^m,\vz\neq\textbf{0}} |\frac{\vert W\vz\vert}{\vert \vz\vert}|$ which is equivalent to force each $W_l$ to be an orthogonal matrix. The orthogonal constraint allows simple calculation of the inverse of $W_{l}$.

\paragraph{Padding loss.}
To force sparsity from the second to $(L-1)$-th layers, we add a zero padding loss to each of these layers. For the $l$-th layer whose target dimension is $s_{l}$, pad the last $m-s_{l}$ elements of $\vz^{(l+1)}$ with zeros and panish these elements with $L_{1}$ norm loss:
\begin{align}
    L_{pad} = \sum_{l=2}^{L-1} \beta^{(l)} \sum_{i=s^{(l)}}^{m} \vert \vz^{(l+1)}_i\vert,
    \label{Eq.CH_3.3-pad}
\end{align}
where $\{\beta^{(l)}\}$ are loss weights. The target dimension $s_{l}$ can be set heuristically.

\paragraph{Linear transformation head.}
We use the linear transformation head to achieve the linear compression step in our NLDR process, which is a transformation between the orthogonal basis of high dimension and lower dimension. Thus, we apply the row orthogonal constraint to $W_{L}$.

\paragraph{LIS loss.}
Since the linear DR is applied at the end of the NLDR process, we apply \textit{locally isometric smoothness} (LIS) constraint \cite{2020-MLDL} to preserve the local geometric properties. Take the LIS loss in the $l$-th layer as an example:
\begin{align}
	L_{LIS} = \sum_{i=1}^{n}\sum_{j \in \mathcal{N}_{{i}}^k} \left\|d_{X}(\vx_{i}, \vx_{j}) - d_{Z}(\vz^{(l)}_{i}, \vz^{(l)}_{j})\right\|,
	\label{Eq.CH_3.3-LIS}
\end{align}
where $\mathcal{N}_{{i}}^k$ is a set of ${x_i}$'s $k$-nearest neighborhood in the input space, and $d_{X}$ and $d_{Z}$ are the distance of the input and the latent space, which can be approximated by Euclidean distance in local open sets.

\paragraph{Push-away loss.}
In the real case discussed in Section \ref{CH_3.3}, the latent space of the $(L-1)$-th layer can approximately be a hyperplane in Euclidean space, so that we introduce push-away loss to repel the non-adjacent sample points of each $x_{i}$ in its $B$-radius neighborhood in the latent space. It deflates the manifold locally when acting together with $L_{LIS}$ in the linear DR. Similarly, $L_{push}$ is applied after the linear transformation in the $l$-th layer:
\begin{align}
	L_{push} = - \sum_{i=1}^{n}\sum_{j \in \mathcal{N}_{{i}}^k}\1_{d_{Z}(\vz_{i}^{(l)}, \vz_{j}^{(l)})<B} \log \left( 1+ d_{Z}(\vz_{i}^{(l)}, \vz_{j}^{(l)})\right),
	\label{Eq.CH_3.3-pushaway}
\end{align}
where $\1(.)\in\{0,1\}$ is the indicator function for the bound of $B$.

\paragraph{Extra head.}
In order to force the first $L-1$ layers of the network to achieve NLDR gradually, we introduce auxiliary DR branches, called \textit{extra heads}, after the second layer to the $(L-1)$-th layer. The structure of each \textit{extra head} is the same as the linear transformation head and will be discarded after training. $L_{extra}$ is written as
\begin{align}
	L_{extra} = \sum_{l=1}^{L-1} \gamma^{(l)}(L_{LIS} + \mu^{(l)}L_{push}),
	\label{Eq.CH_3.3-extra}
\end{align}
where $\{\gamma^{(l)}\}$ and $\{\mu^{(l)}\}$ are loss weights which can be set based on $\{s_l\}$.

\paragraph{Inverse process.}
The inverse process is the decoder directly obtained by the first $L-1$ layers of the encoder given by Eq. (\ref{Eq.CH_3.3-1}), which is not involved in the training process. When the target dimension $s'$ is equal to $s$, the inverse of the layer-$L$ can be solved by some existing methods such as compressed sensing or eigenvalue decomposition.

\section{Experiment}

In this section, we first evaluate the proposed \textit{inv-ML} achieved by \textit{i-ML-Enc} in Section \ref{CH_4.1}, then investigate the property of data manifolds with \textit{i-ML-Enc} in Section \ref{CH_4.2}. The properties of \textit{i-ML-Enc} are further studied in Section \ref{CH_4.3}. We carry out experiments on \textbf{seven datasets}: (\romannumeral1) Swiss roll \cite{2011-JMLR-sklearn}, (\romannumeral2) Spheres \cite{2020-ICML-TopoAE} and Half Spheres, (\romannumeral3) USPS \cite{1994-USPS}, (\romannumeral4) MNIST \cite{1998-MNIST}, (\romannumeral5) KMNIST \cite{2018-KMNIST}, (\romannumeral6) FMNIST \cite{2017-FMNIST}, (\romannumeral7) COIL-20 \cite{1996-coil20}. The first two datasets are uniformly sampled on synthetic manifolds which can reflect mathematical properties of NLDR. The later five are real-world datasets where samples lie on circular manifolds (COIL-20) and cluster manifolds (MNIST, USPS, KMNIST, FMNIST). 
The following settings of \textit{i-ML-Enc} are used for all datasets: LeakyReLU with $\alpha=0.1$; Adam optimizer \cite{2015-ICLR-Adam} with learning rate $lr=0.001$ for $8000$ epochs; the local neighborhood is determined by kNN with $k=15$. The implementation is based on the PyTorch 1.3.0 library running on NVIDIA v100 GPU, and the source code is available at \url{https://github.com/Westlake-AI/inv-ML}.

\renewcommand\tabcolsep{5.0pt}
\begin{table}[htb]
	\setlength{ \abovecaptionskip}{0.cm}
	\caption{Comparison in NLDR, invertible and generalization qualities on MNIST and COIL-20.}
	\label{table:CH_4_1_SOTA}
    \begin{center}
    \resizebox{0.96\columnwidth}{!}{
		\begin{tabular}{c|lllllll}
			\toprule
			\multicolumn{1}{l}{}
			& Algorithm      & RMSE            & MNE             & Trust           & Cont            & $l$-MSE        & Acc             \\
			\hline
			\multirow{11}{*}{\rotatebox{90}{MNIST}}
			& MLLE           & -               & -               & 0.6709          & 0.6573          & 36.80          & 0.8341          \\
			& t-SNE          & -               & -               & 0.9896          & 0.9886          & 48.07          & 0.9246          \\
			& ML-Enc         & -               & -               & 0.9862          & \textbf{0.9927} & 18.98          & 0.9326          \\
			& VAE            & 0.5263          & 33.17           & 0.9712          & 0.9703          & 22.79          & 0.8652          \\
			& GRAE           & 0.4324          & 17.32           & 0.9811          & 0.9796          & 20.45          & 0.8769          \\
			& TopoAE         & 0.5178          & 31.45           & \textbf{0.9915} & 0.9878          & 24.98          & 0.8993          \\
			& ML-AE          & 0.4012          & 16.84           & 0.9893          & 0.9926          & 19.05          & \textbf{0.9340} \\
			& i-ML-Enc (L)   & \textbf{0.0457} & \textbf{0.5085} & 0.9906          & 0.9912          & \textbf{18.16} & 0.9316          \\
			\cline{2-8}
			& INN            & 0.0615          & 0.5384          & 0.9851          & 0.9823          & 7.494          & 0.9176          \\
			& i-RevNet       & 0.0443          & \textbf{0.4679} & 0.9118          & 0.8785          & 6.958          & - \\
			& i-ResNet       & 0.0502          & 0.6422          & 0.9149          & 0.8922          & 10.78          & - \\
			& i-ML-Enc(L-1)  & \textbf{0.0407} & 0.5085          & \textbf{0.9986} & \textbf{0.9973} & \textbf{5.895} & \textbf{0.9580} \\
			\hline
			\multirow{11}{*}{\rotatebox{90}{COIL-20}}
			& t-SNE         & -               & -               & 0.9911          & \textbf{0.9954} & 17.22          & 0.9039          \\
			& ML-Enc        & -               & -               & 0.9920          & 0.9889          & \textbf{9.961} & \textbf{0.9564} \\
			& AE            & 0.3507          & 24.09           & 0.9745          & 0.9413          & 11.45          & 0.8958          \\
			& GRAE          & 0.2685          & 23.57           & 0.9840          & 0.9705          & 25.36          & 0.8912          \\
			& TopoAE        & 0.4712          & 26.66           & 0.9768          & 0.9625          & 27.19          & 0.9043          \\
			& ML-AE         & 0.1220          & 16.86           & 0.9914          & 0.9885          & 10.34          & 0.9548          \\
			& i-ML-Enc (L)  & \textbf{0.0312} & \textbf{1.026}  & \textbf{0.9921} & 0.9871          & 11.13          & 0.9386          \\ \cline{2-8} 
			& INN           & 0.0758          & 0.8075          & 0.9791          & 0.9681          & 8.595          & 0.9936          \\
			& i-RevNet      & 0.0508          & 0.7544          & 0.9316          & 0.9278          & 9.803          & -  \\
			& i-ResNet      & 0.0544          & \textbf{0.7391} & 0.9258          & 0.9136          & 10.41          & -  \\
			& i-ML-Enc(L-1) & \textbf{0.0312} & 0.9263          & \textbf{0.9940} & \textbf{0.9937} & \textbf{7.539} & \textbf{1.000}  \\
			\bottomrule
		\end{tabular}}
	\end{center}
\end{table}

\begin{figure*}[!htb]
  \centering
  \includegraphics[width=4.9in]{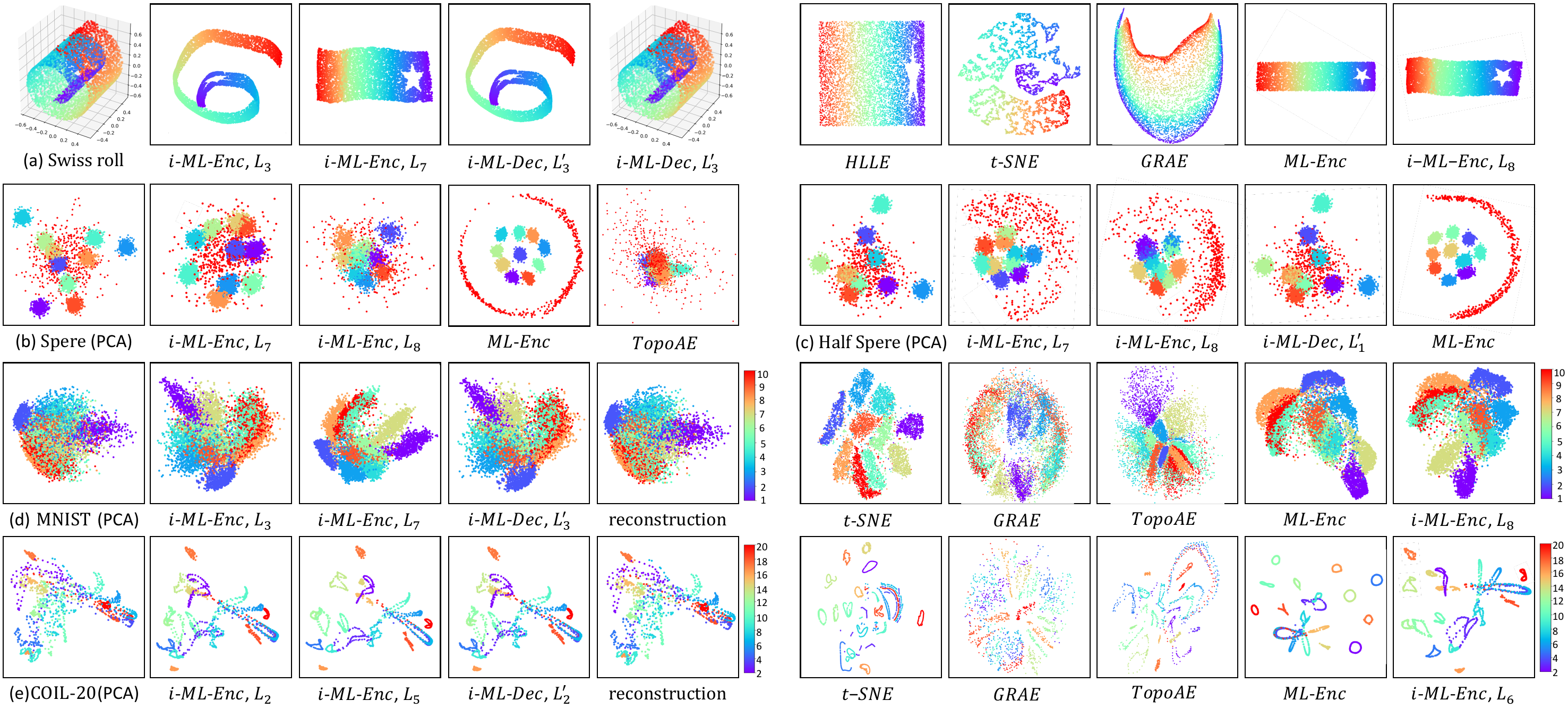}
  \caption{Visualization of NLDR results of \textit{i-ML-Enc} and relevant methods. All the high-dimensional results are visualized by PCA and the target dimension $s'=2$. (a) shows NLDR and its inverse process of \textit{i-ML-Enc} on the test set of Swiss roll in the case of $d=s=s'$. We show the cases of $s'<d\le s$ and $s'=d\le s$ by comparing (b)(c): (b) shows the failure case of reducing spheres $S^{100}$ sampled in $\mathbb{R}^{101}$ into $10$-D, while (c) shows results of reducing half-spheres $S^{10}$ sampled in $\mathbb{R}^{101}$ into $10$-D. In (b), TopoAE preserves the topological structure of those hyper-spheres. ML-Enc only maintains the geometric structure of circles but collapses into bad topological structures. In both cases, \textit{i-ML-Enc} maintains the same topology as the input data in the first $7$ layers, though it fails to achieve NLDR in (b). (d) and (e) show results of two sparse cases on MNIST and COIL-20: The left columns provide the invertible NLDR process of \textit{i-ML-Enc} which are homeomorphic mappings. Because of the clustering effect, it is vital to focus on the local geometric structure while simply preserving the correct relationship between sub-manifolds. The results of ML-Enc and t-SNE show clear cluster structures and geometric structures of sub-manifolds. GRAE and TopoAE show more mixed results because of their over-reliance on topological structures. The results of \textit{i-ML-Enc} provide similar local structures shapes as ML-Enc, but more connection between clusters.
  }
  \label{figure:CH_4_1_SOTA}
\end{figure*}

\subsection{Methods Comparison}
\label{CH_4.1}
To verify the invertible NLDR ability of \textit{i-ML-Enc} and analyze different cases of NLDR, we compare it with eight typical methods in NLDR and inverse scenarios on both synthetic (Swiss roll, Spheres and Half Spheres) and real-world datasets (USPS, MNIST, FMNIST and COIL-20). \textbf{Eight methods for manifold learning}: Isomap \cite{2000-science-Isomap}, MLLE \cite{2007-MLLE}, t-SNE \cite{2008-JMLR-tSNE} and ML-Enc \cite{2020-MLDL} are compared for NLDR; four AE-based methods VAE \cite{2014-ICLR-VAE}, GRAE \cite{2020-GRAE}, TopoAE \cite{2020-ICML-TopoAE} and ML-AE \cite{2020-MLDL} are compared for reconstructible manifold learning. \textbf{Three methods for inverse models}: INN \cite{2019-GCPR-auto_INN}, i-RevNet \cite{2018-ICLR-iRevNet}, and i-ResNet \cite{2019-ICML-iResNet} are compared for bijective inverse property. Among them, i-RevNet and i-ResNet are supervised algorithms while the rest are unsupervised. For a fair comparison in this experiment, we adopt $8$ layers neural network for all the network-based methods except i-RevNet and i-ResNet. {\bf Hyperparameter} values of \textit{i-ML-Enc} and configurations of datasets are provided in \textbf{Appendix \ref{A_2}}.

\paragraph{Evalution metrics.}
We evaluate an invertible NLDR algorithm from three aspects: (\romannumeral1) Invertible property. Reconstruction MSE (\textbf{RMSE}) and maximum norm error (\textbf{MNE}) measure the difference between the input data and reconstruction results by norm-based errors. (\romannumeral2) NLDR quality. Trustworthiness (\textbf{Trust}), Continuity (\textbf{Cont}) \cite{2006-lMDS}, and latent MSE (\textbf{l-MSE}) \cite{2020-MLDL} are used to evaluate the quality of the low-dimensional representation. (\romannumeral3) Generalization ability. Mean accuracy (\textbf{Acc}) of linear classification on the learned representation measures models' generalization ability to downstream tasks. Their exact definitions and purpose are given in \textbf{Appendix \ref{A_1}}.

\paragraph{Comparison and Conclusion.}
Tab. \ref{table:CH_4_1_SOTA} compares the \textit{i-ML-Enc} with the relevant methods on MNIST and FMNIST datasets, more results and detailed analysis on other datasets are given in \textbf{Appendix \ref{A_2}}. 
The process of invertible NLDR of \textit{i-ML-Enc} and comparing results of typical methods are visualized in Fig. \ref{figure:CH_4_1_SOTA}. We can conclude: 
(\romannumeral1) \textit{i-ML-Enc} achieves invertible NLDR in the first stage with great NLDR and generalization qualities. The representation in the $L-1$-th layer of \textit{i-ML-Enc} mostly outperforms all comparing methods for both invertible and NLDR metrics without losing information of the data, while other methods drop geometric and topological information to some extent. (\romannumeral2) \textit{i-ML-Enc} tries to keep more geometric and topological structure in the second stage in the case of $s^{'}<d\le s$. The $L$-th layer of \textit{i-ML-Enc} shows high consistency with its $L-1$-th layer and comparable NLDR performance in visualization results. (\romannumeral3) \textit{i-ML-Enc} provides more reliable and explainable representations of the data manifold because of its good mathematic properties.

\subsection{Latent Space Interpolation}
\label{CH_4.2}
Since the first stage of \textit{i-ML-Enc} is nearly homeomorphism, we carry out linear interpolation experiments in both the input space and the $(L-1)$-th layer latent space to analyze the intrinsic continuous manifold and verify the latent results by its inverse process. A good low-dimensional representation of the manifold should not only preserve the local properties, but also be flatter (with lower curvature) than the high-dimensional input space. Thus, we expect that the local linear interpolation results in the latent space should be more reliable than in the input space.

\paragraph{Interpolation datasets.}
The manifold learning difficulties of five datasets can be roughly analyzed in terms of \textbf{sampling ratio}, \textbf{image entropy}, \textbf{texture}, and performances on \textbf{classification tasks}: 
(\romannumeral1) Sampling ratio. The input dimension and sample number reflect the sampling ratio. In the case of sufficient sampling, the sample number nearly has an exponential relationship with the input dimension. Thus, the sampling ratio of USPS is higher than others. (\romannumeral2) Image entropy. The Shannon entropy of the histogram measures the information content of images. It shows that USPS has richer grayscale than MNIST(256). The information content of MNIST(784), KMNIST, and FMNIST shows an increasing trend. (\romannumeral3) Texture. The standard deviation (std) of the histogram reflects the texture information in images. (\romannumeral4) Classification tasks. Performances of kNN classifier \cite{2011-JMLR-sklearn} on the input space reflect the credibility of the neighborhood system. The credibility decreases gradually from USPS, MNIST, KMNIST to FMNIST. 
In a nutshell, we can conclude that the complexity of data manifolds increases from USPS(256), MNIST(256), MNIST(784), KMNIST(784) to FMNIST(784).

\begin{figure}[!htb]
  \centering
  \includegraphics[width=4.95in]{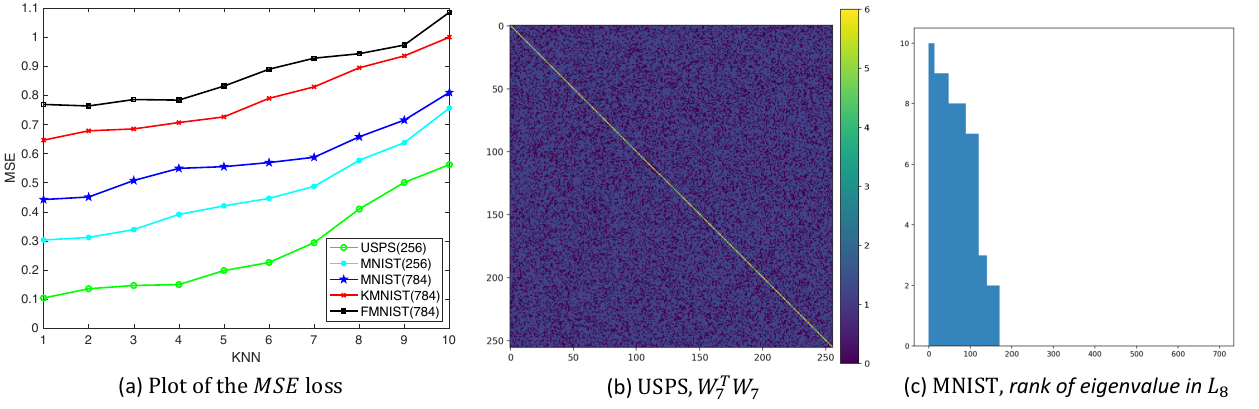}
  \caption{(a) shows the MSE loss of $1$ to $10$ nearest neighbors interpolation results on five datasets. It reflects the reliability of linear approximation in different low-dimensional representations. (b) shows the orthogonality of the weight matrix $W_{7}$ in \textit{inv-ML-Enc} trained on USPS (256x256) dataset. The elements are ranged from $10^0$ to $10^5$ after min-max normalization and rescaling, indicating that $W_{7}$ is nearly an orthogonal matrix. (c) shows the rank of eigenvalues (by SVD) of the $8$-th layer output of \textit{i-ML-Enc} on MNIST test set, which range from $10^0$ to $10^10$ by rescaling. The matrix rank of the output is $125$, and the extra $46$-D can be regarded as some machine errors when performs PCA.}
  \label{figure:ch4_2_0}
\end{figure}

\paragraph{K-nearest neighbor interpolation.}
We verify the reliability of the low-dimensional representation in a small local system by kNN interpolation. Given a sample $\vx_{i}$, randomly select $\vx_{j}$ in $\vx_{i}$'s k-nearest neighborhood in the latent space to form a sample pair $(\vx_{i},\vx_{j})$. Perform linear interpolation of the latent representation of the pair and get reconstruction results for evaluation as:
	$\hat \vx_{i,j}^{t} = \psi^{-1}( t\psi(\vx_{i}) + (1-t)\psi(\vx_{j})),\ t\in[0,1].$
The experiment is performed on \textit{i-ML-Enc} with $L=6$ and $K=15$, training with $9298$ samples for USPS and MNIST(256), $20000$ sapmles for MNIST(784), KMNIST, FMNIST. 
We evaluate kNN interpolation from two aspects: 
(\romannumeral1) Calculate the MSE loss between reconstruction results of the latent interpolation $\hat \vx^{t}_{i,j}$ and the corresponding input interpolation results $\vx^t_{i,j} = t\vx_i + (1-t)\vx_j$. A larger MSE loss indicates the worse fitting to the data manifold. Notice that this MSE loss is only a rough measurement of kNN interpolation when $k$ is small. Fig. \ref{figure:ch4_2_0} shows evaluation results with $k=1,2,...,10$. (\romannumeral2) Visualize typical results of the input space and the latent space for comparison, as shown in Fig. \ref{figure:ch4_2_1}. More results and analysis are given in \textbf{Appendix \ref{A_3.2}}.
We further employ \textit{geodesic interpolation} between two distant samples pairs in the latent space to analyze topological structures. Given a sample pair $(x_{i},x_{j})$ from different clusters, we select the three intermediate sample pairs $(x_{i},x_{i_1})$, $(x_{i_1},x_{i_2})$, $(x_{i_2},x_{j})$ with $k\le20$ along the geodesic path in latent space. Visualization results are given in \textbf{Appendix \ref{A_3.2}}.
The latent results show no overlap of multiple submanifolds in the geodesic path.

\begin{figure}[!htb]
	\centering
	\includegraphics[width=4.7in]{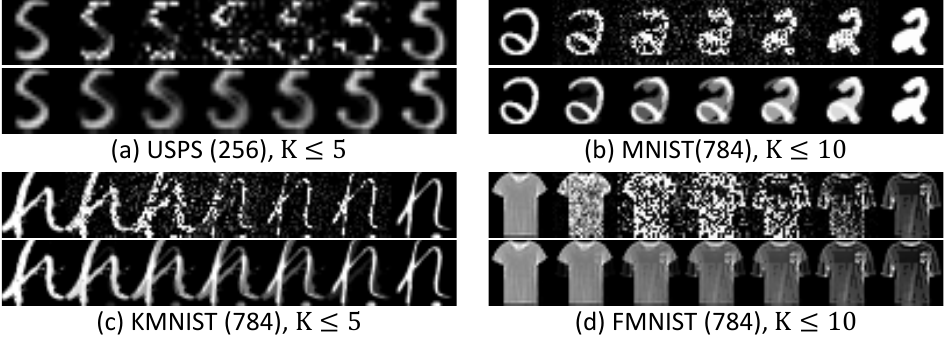}
	\caption{Results of kNN interpolation. For each dataset, the upper and lower rows show latent space and input space results respectively. From a overall aspect, the latent results show more noise because \textit{inv-ML-Enc} is not a AE-based or generative model which optimize reconstruction results explicitly. But the latent results are more reliable than the input. For example, left latent interpolation results are similar to the left sample which show less overlapping and pseudo-contour than the input results.}
	\label{figure:ch4_2_1}
\end{figure}

\paragraph{Comparison and Conclusion.}
Compared with results of the kNN and geodesic interpolation, we can conclude: 
(\romannumeral1) Because of the sparsity of the latent space, noises are inevitable on the latent results. Empirically, the reliability of the latent interpolation decreases with the expansion of the local neighborhood on the same dataset. 
(\romannumeral2) The latent results of kNN interpolation get worse in the following cases: for similar manifolds, when the sampling rate is lower (indicated by USPS(256), MNIST(256) and MNIST(784)); with the same sampling rate, the manifold becomes more complex (indicated by MNIST(784), KMNIST to FMNIST). They indicate that the confidence of the tangent space estimated by local neighborhood decreases on more complex manifolds with sparse sampling. 
(\romannumeral3) The interpolation between two samples in latent space is smoother than that in the input space, validating the flatness and density of the lower-dimensional representation learned by \textit{i-ML-Enc}. 
Overall, we infer that the unreliable approximation of the local tangent space by the local neighborhood is the basic reason for the manifold learning fails in the real-world case, because the geometry should be preserved in the first place. To come up with this common situation, it is necessary to import other prior assumptions or knowledge when the sampling rate of data manifolds is relatively low, e.g., the Euclidean space assumption, semantic information of down-stream tasks.

\subsection{Analysis}
\label{CH_4.3}
\paragraph{Analysis on loss terms.}
We perform an ablation study to evaluate the effects of the proposed network structure and loss terms in \textit{i-ML-Enc} on MNIST, USPS, KMNIST, FMNIST, and COIL-20. Based on ML-Enc, three proposed parts are added: the \textit{extra head} (\textbf{Ex}), the orthogonal loss $\mathcal{L}_{orth}$ (\textbf{Orth}), the padding loss $\mathcal{L}_{pad}$ (\textbf{Pad}). Besides the previous six indicators, we introduce the rank of the output matrix of the layer $L-1$ as $r(Z^{L-1})$, to measure the sparsity of the high-dimensional representation. We conclude that the combination \textbf{Ex+Orth+Pad} is the best to achieve invertible NLDR of $s$-sparse by a series of equidimensional layers. The detailed analysis of experimental results is given in \textbf{Appendix A.4}.

\paragraph{Orthogonality and sparsity.}
We further discuss the orthogonality of weight matrices and $s$-sparse representations in the first stage of \textit{i-ML-Enc}. We find that the first $L-1$ layers of \textit{i-ML-Enc} are nearly strict orthogonal mappings because each layer satisfies $\vert| W_{l}^TW_{l}-I\vert|<10^{-5}$, as illustrated in Fig. \ref{figure:ch4_2_0} (b). Meanwhile, the $L-1$-th layer output of \textit{i-ML-Enc} achieves sparsity. Taking the $8$-th layer output of \textit{i-ML-Enc} on MNIST test set as an example, as shown in Fig. \ref{figure:ch4_2_0} (c). We can construct a $125$-D linear subspace with $125$ orthogonal base vectors decomposed from the output matrix and reconstruct to the original space ($784$-D) without losing information by PCA \cite{2011-JMLR-sklearn} and its inverse transform. It indicates a low-dimensional constrain is learned by \textit{inv-ML-Enc}. Thus, we conclude that an invertible NLDR of data manifolds can be learned by \textit{i-ML-Enc} in the \textit{sparse coordinate transformation}.

\paragraph{Relationship between $s$-sparse and intrinsic dimension $d$.}
We notice that the $s$-sparse achieved by the first stage of \textit{i-ML-Enc} is higher than the approximate intrinsic dimension $d$ on each dataset, e.g. $116$-sparse on USPS and $125$-sparse on MNIST. We found the following reasons: 
(\romannumeral1) Because the data manifolds are usually quite complex but sampling sparsely, the lowest isometric embedding dimension is between $d$ to $2d$ according to Nash Embedding Theorem and the hyper-plane hypothesis. The $s$ obtained by \textit{i-ML-Enc} on each dataset is nearly in the interval of $[d, 2d]$, which is not the true intrinsic dimension of the manifolds. 
(\romannumeral2) The proposed \textit{i-ML-Enc} is not optimized enough, which serves as a simple network implementation of inv-ML. We need to design a better implementation model if we want to approach the lower embedding dimension to preserve both geometry and topology.

\renewcommand\tabcolsep{4.0pt}
\begin{table}[htb]
	\setlength{ \abovecaptionskip}{0.cm}
	\caption{Ablation study of proposed loss terms in \textit{i-ML-Enc} on MNIST.}
	\label{table:CH_4_3_Ablation}
    \begin{center}
    \resizebox{0.94\columnwidth}{!}{
		\begin{tabular}{c|lllllll}
		\toprule
		\multicolumn{1}{l}{}
        &              & RMSE            & MNE             & Trust           & Cont            & Acc             & $r(Z^{L-1})$ \\
        \hline
        \multirow{6}{*}{\rotatebox{90}{MNIST}}
        & ML-AE        & 0.4012          & 16.84           & 0.9893          & 0.9926          & \textbf{0.9340} & 15          \\
        & ML-Enc       & -               & -               & 0.9862          & \textbf{0.9927} & 0.9326          & 14          \\
        & +Ex          & -               & -               & 0.9891          & 0.9812          & 0.9316          & \textbf{12} \\
        & +Orth        & \textbf{0.0056} & \textbf{0.1275} & 0.9652          & 0.9578          & 0.8807          & 716         \\
        & +Ex+Orth     & 0.0341          & 0.4255          & 0.9874          & \textbf{0.9927} & 0.9298          & 361         \\
        & +Ex+Orth+Pad & 0.0457          & 0.5085          & \textbf{0.9906} & 0.9912          & 0.9316          & 125         \\
		\bottomrule
		\end{tabular}}
	\end{center}
\end{table}

\section{Conclusion}
To fill the gap between theoretical and real-world applications of manifold-based DR, we introduce a novel invertible NLDR process \textit{inv-ML} and a neural network implementation \textit{inv-ML-Enc} to verify the proposed process. 
Firstly, the \textit{sparse coordinate transformation} is learned to find a flatter and denser low-dimensional representation with preservation of geometry and topology of data manifolds. Secondly, we discuss the condition of NLDR and information loss with different target dimensions in \textit{linear compression}. Experiment results of \textit{i-ML-Enc} on seven datasets validate the proposed invertible NLDR process and the sparsity of learned low-dimensional representations. Further, the interpolation experiments reveal that finding a reliable tangent space by the local neighborhood on real-world datasets is the inherent defect of manifold-based DR methods.

\section*{Acknowledgments}
This work was performed during the internship of Siyuan Li and Haitao Lin at Westlake University. We thank Di Wu for helpful insights on hyperparameters tuning and polishing the writing.

%
%
%
\bibliographystyle{splncs04}
\bibliography{inv_ML}

%
\clearpage

\setcounter{table}{0}
\setcounter{figure}{0}
\renewcommand{\thetable}{A\arabic{table}}
\renewcommand{\thefigure}{A\arabic{figure}}

\appendix
\section{Appendix}

\subsection{Definitions of Performance Metrics}
\label{A_1}
As for NLDR tasks, We adopt the performance metrics used in MLDL \cite{2020-MLDL} and TopoAE \cite{2020-ICML-TopoAE} to measure topology-based manifold learning, and add a new indicator to evaluate the generalization ability of the latent space. Essentially, the related indicators are defined based on comparisons of the local neighborhood of the input space and the latent representation. As for the invertible property, we adopted the norm-based reconstruction metrics, i.e. the $L_2$ and $L_{\infty}$ norm errors, which are based on the inputs. The following notations are used in the definitions: 
$d^{(l)}_{i,j}$ is the pairwise distance in space $Z^{(l)}$; $\mathcal{N}_{i,k}^{(l)}$ is the set of indices to the $k$-nearest neighbors ($k$-NN) of $z^{(l)}_i$ in latent space, and $\mathcal{N}_{i,k}$ is the set of indices to the $k$-NN of $x_i$ in input space; $r^{(l)}_{i,j}$ is the closeness rank of $z^{(l)}_j$ in the $k$-NN of $z^{(l)}_i$. 
The evaluation metrics are defined below:\\
\begin{enumerate}
    \item[(1)] \textbf{RMSE} (invertible quality). This indicator is commonly used to measure reconstruction quality. Based on the input $x$ and the reconstruction output $\hat x$, the mean square error (MSE) of the $L_2$ norm is defined as:
    \begin{align*}
       RMSE = (\frac{1}{N^2} \sum_{i=1}^{N} (\vx_i - \vz_i)^2 )^{\frac{1}{2}}.
    \end{align*}
    \item[(2)] \textbf{MNE} (invertible quality). This indicator is designed to evaluate the bijective property of a $L$ layers neural network model. Specifically, taking each invertible unit in the network, calculate the $L_{\infty}$ norm error of the input and reconstruction output of each corresponding layer, and choose the maximum value among all units. If a model is bijective, this indicator can reflect the stability of the model:
    \begin{align*}
       MNE = \max_{1\leq l\leq L-1}\Vert \vz_l - \hat \vz_l \Vert_{\infty},\ l=1,2,...L.
    \end{align*}
    \item[(3)] \textbf{Trust} (embedding quality). This indicator measures how well neighbors are preserved between the two spaces. The $k$ nearest neighbors of a point are preserved when going from the input space $X$ to space $Z^{(l)}$:
    \begin{align*}
        Trust=\frac{1}{k_2-k_1+1} \sum_{k=k_{1}}^{k_{2}} \left\{1-\frac{2}{Mk (2 M-3 k-1)} \sum_{i=1}^{M} \sum_{j \in \mathcal{N}_{i,k}^{(l)},j \not\in \mathcal{N}_{i,k}}(r^{(l)}_{i,j}-k)\right\}
    \end{align*}
    where $k_1$ and $k_2$ are the bounds of the number of nearest neighbors, so averaged for different $k$-NN numbers. 
    \item[(4)] \textbf{Cont} (embedding quality). This indicator is asymmetric to \textbf{Trust}. It checks to what extent neighbors are preserved from the latent space $Z^{(l)}$ to the input space $X$:
    \begin{align*}
        Cont=\frac{1}{k_2-k_1+1} \sum_{k=k_{1}}^{k_{2}}  \left\{1-\frac{2}{Mk  (2 M-3 k-1)} \sum_{i=1}^{M} \sum_{j \in \mathcal{N}_{i,k},j \not\in \mathcal{N}_{i,k}^{(l)}}(r^{(l)}_{i,j}-k)\right\}
    \end{align*}
    \item [(5)] \textbf{$K$min} and \textbf{$K$max} (embedding quality). Those indicators are the minimum and maximum of the local bi-Lipschitz constant for the homeomorphism between input space and the $l$-th layer, with respect to the given neighborhood system:
	\begin{align*}
		K_{\rm min}=\min_{1\leq i \leq M} \max _{j \in \mathcal{N}_{i,k}^{(l)} } K_{i, j},\ 
		K_{\rm max}=\max_{1\leq i \leq M} \max _{j \in \mathcal{N}_{i,k}^{(l)} } K_{i, j},
	\end{align*}
    where $k$ is that for $k$-NN used in defining $N_i$ and
    \begin{align*}
        K_{i, j}=\max \left\{\frac{d^{(l)}_{i,j}}{d^{(l')}_{i,j}}, \frac{d^{(l')}_{i,j}}{d^{(l)}_{i,j}}\right\}.
	\end{align*}
	\item [(6)] \textbf{$l$-MSE} (embedding quality). This indicator is to evaluate the distance disturbance between the input space and latent space with $L_{2}$ norm-based error.
	\begin{align*}
		{lMSE} = (\frac{1}{N^2} \sum_{i=1}^{N} \sum_{j=1}^{N}\left \Vert d_X(\vx_i,\vx_j)-d_Z(h (\vx_i),h (\vx_j)) \right \Vert )^{\frac{1}{2}}.
	\end{align*}
	\item [(7)] \textbf{ACC} (generalization ability). In general, a good representation should have a good generation ability to downstream tasks. To measure this ability, logistic regression \cite{2011-JMLR-sklearn} is performed after the learned latent representation. We report the mean accuracy on the test set for $10$-fold cross-validation.
\end{enumerate}

\subsection{Method Comparison}
\label{A_2}
\textbf{Configurations of datasets.} \quad
The NLDR performance and its inverse process are verified on both synthetic and real-world datasets. As shown in Tab. \ref{table:A_2_introduction}, we list the \textbf{type} of the dataset, the \textbf{class} number of clusters, the \textbf{input} dimension $m$, the \textbf{target} dimension $s'$, the \textbf{intrinsic} dimension $d$ which is only an approximation for the real-world dataset, the number of \textbf{train} and \textbf{test} samples, and the \textbf{logistic} classification performance on the raw input space. Among them, Swiss roll serves as an ideal example of information-lossless NLDR; Spheres, whose target dimension $s^{'}$ is lower than the intrinsic dimension $s$, serves as an extreme case of NLDR compared to Half-spheres; four image datasets with increasing difficulties are used to analyze complex situations in real-world scenarios. Additionally, the lower bound and upper bound of the intrinsic dimension of real-world datasets are approximated by \cite{2005-ICML-instinct} and AE-based INN \cite{2019-GCPR-auto_INN}. Specifically, the upper bound can be found by the grid search of different bottlenecks of the INN, and we report the bottleneck size of each dataset when the reconstruction MSE loss is almost unchanged.
\renewcommand\tabcolsep{5.0pt}
\begin{table}[htb]
    \setlength{\abovecaptionskip}{0.cm}
	\caption{Brief introduction to the configuration of datasets for method comparison.}
	\label{table:A_2_introduction}
	\begin{center}
	    \resizebox{0.98\columnwidth}{!}{
		\begin{tabular}{lllllllll}
			\toprule
			Dataset      & Type       & Class & Input $m$ & Target $s'$ & intrinsic $d$ & Train  & Test  & Logistic \\
			\hline
			Swiss roll   & synthetic  & -     & 3         & 2           & 2            & 800    & 8000  & -        \\
			Spheres      & synthetic  & -     & 101       & 10          & 101          & 5500   & 5500  & -        \\
			Half-spheres & synthetic  & -     & 101       & 10          & 10           & 5500   & 5500  & -        \\
			USPS         & real-world & 10    & 256       & 10          & 10 to 80     & 4649   & 4649  & 0.9381   \\
			MNIST        & real-world & 10    & 784       & 10          & 10 to 100    & 20000  & 10000 & 0.8943   \\
			FMNIST       & real-world & 10    & 784       & 10          & 20 to 140    & 20000  & 10000 & 0.7984   \\
			COIL-20      & real-world & 20    & 4096      & 20          & 20 to 260    & 1440   & 1440  & 0.9974   \\
			\bottomrule
		\end{tabular}}
	\end{center}
\end{table}

\textbf{Hyperparameter values.} \quad
Basically, \textit{i-ML-Enc} is trained with Adam optimizer \cite{2015-ICLR-Adam} and learning rate $lr=0.001$ for $8000$ epochs. We set the layer number $L=8$ for most datasets but $L=6$ for COIL-20. The bound in push-away loss is set $B=3$ in most datasets but removed in Spheres and Half-spheres. 
We set the hyperparameter based on two intuitions: (1) the implementation of \textit{sparse coordinate transformation} should achieve DR on the premise of maintaining homeomorphism; (2) NLDR should be achieved gradually from the first to $(L-1)$-th layer because NLDR is impossible to achieve by a single nonlinear layer. Based on (1), we decrease the \textit{extra heads} weights $\gamma$ linearly for epochs from $2000$ to $8000$, while linearly increase the orthogonal loss weights $\alpha$ for epochs from $500$ to $2000$. Based on (2), we approximate the DR trend by exponential series. For the \textit{extra heads}, the target dimension decrease exponentially from $m$ to $s^{'}$ for the $2$-th to $(L-1)$-th layer, and the push-away loss weights $\mu$ increase linearly. Similarly, the padding weight $\beta$ should increase linearly. Because the intrinsic dimension is different from each real-world dataset, we adjust the prior hyperparameters according to the approximated intrinsic dimension.

\renewcommand\tabcolsep{5.0pt}
\begin{table}[htb]
    \setlength{\abovecaptionskip}{0.cm}
	\caption{Comparison in embedding and invertible quality on Swiss roll and Half-spheres.}
	\label{table:A_2_toy}
	\begin{center}
	    \resizebox{0.98\columnwidth}{!}{
		\begin{tabular}{l|llllllll}
			\toprule
			\multicolumn{1}{l}{Dataset}
			& Algorithm      & RMSE            & MNE             & Trust           & Cont            & $K$min         & $K$max         & $l$-MSE         \\
			\hline
			\multirow{9}{*}{Swiss Roll}
			& Isomap         & -               & -               & 0.9834          & 0.9812          & 1.213          & 43.55          & 0.0756          \\
			& t-SNE          & -               & -               & 0.9987          & 0.9843          & 10.96          & 1097           & 3.407           \\
			& RR             & -               & -               & 0.9286          & 0.9847          & 4.375          & 187.7          & 0.0453          \\
			& ML-Enc         & -               & -               & \textbf{0.9999} & 0.9985          & \textbf{1.000} & \textbf{2.141} & \textbf{0.0039} \\
			& AE             & 0.3976          & 10.55           & 0.8724          & 0.8333          & 1.687          & 4230           & 0.0308          \\
			& VAE            & 0.7944          & 13.97           & 0.5064          & 0.6486          & 1.51           & 4809           & 0.0397          \\
			& TopoAE         & 0.5601          & 11.61           & 0.9198          & 0.9881          & 1.194          & 220.6          & 0.1165          \\
			& ML-AE          & 0.0208          & 8.134           & 0.9998          & 0.9847          & 1.005          & 2.462          & 0.0051          \\
			& i-ML-Enc (L)   & \textbf{0.0048} & \textbf{0.0649} & 0.9996          & \textbf{0.9986} & 1.004          & 2.471          & 0.0043          \\ \hline
			\multirow{8}{*}{Half-spheres}
			& Isomap         & -               & -               & 0.8701          & 0.9172          & 1.845          & 199.3          & 0.4046          \\
			& t-SNE          & -               & -               & \textbf{0.8908} & 0.9278          & 25.33          & 790.9          & 2.6665          \\
			& RR             & -               & -               & 0.8643          & 0.8516          & 3.047          & 201.2          & 0.4789          \\
			& ML-Enc         & -               & -               & 0.8837          & 0.9305          & 1.029          & 46.35          & \textbf{0.0207} \\
			& AE             & 0.7359          & 11.54           & 0.6886          & 0.7069          & 1.763          & 4112           & 0.0937          \\
			& VAE            & 0.8474          & 14.97           & 0.5398          & 0.6197          & 2.361          & 4682           & 0.1205          \\
			& TopoAE         & 0.9174          & 13.68           & 0.8574          & 0.8226          & 1.375          & 154.8          & 0.4342          \\
			& ML-AE          & 0.6339          & 9.492           & 0.8819          & 0.9293          & \textbf{1.025} & 43.17          & 0.0218          \\
			& i-ML-Enc (L)   & \textbf{0.1095} & \textbf{0.7985} & 0.8892          & \textbf{0.9295} & 1.491          & \textbf{41.25} & 0.0463          \\
			\bottomrule
		\end{tabular}}
	\end{center}
\end{table}
\renewcommand\tabcolsep{5.0pt}
\begin{table}[htb]
    \setlength{\abovecaptionskip}{0.cm}
	\caption{Comparison in embedding and invertible quality on USPS, FMNIST, and COIL-20. ML-Enc shows comparable performance for embedding metrics. Based on ML-Enc, \textit{i-ML-Enc} achieves invertible NLDR in the first stage while maintaining a good generalization ability. It also achieves the top embedding performance for the most NLDR metrics in the second stage when $s'<d\le s$.}
	\label{table:A_2_real}
	\begin{center}
	    \resizebox{0.98\columnwidth}{!}{
		\begin{tabular}{c|lllllllll}
			\toprule
			\multicolumn{1}{l}{Dataset}
			& Algorithm     & RMSE            & MNE             & Trust           & Cont            & $K$min         & $K$max         & $l$-MSE        & Acc             \\
			\hline
			\multirow{10}{*}{USPS}
			& t-SNE         & -               & -               & 0.9831          & 0.9889          & 3.238          & 194.8          & 35.53          & 0.9522          \\
			& ML-Enc        & -               & -               & 0.9874          & 0.9897          & 1.562          & \textbf{52.14} & \textbf{14.88} & 0.9534          \\
			& AE            & 0.6201          & 29.09           & 0.9845          & 0.974           & 4.728          & 87.41          & 17.41          & 0.8952          \\
			& TopoAE        & 0.647           & 30.19           & 0.9830          & 0.9852          & 3.598          & 126.2          & 19.98          & 0.8876          \\
			& ML-AE         & 0.4912          & 11.84           & 0.9879          & \textbf{0.9905} & 1.529          & 55.32          & 15.05          & \textbf{0.9576} \\
			& i-ML-Enc (L)  & \textbf{0.0253} & \textbf{0.3058} & \textbf{0.9886} & 0.9861          & \textbf{1.487} & 60.79          & 15.16          & 0.9435          \\ \cline{2-10}
			& INN           & 0.0535          & 0.5239          & 0.9872          & 0.9843          & 1.795          & 26.38          & 9.581          & 0.9305          \\
			& i-RevNet      & 0.0337          & 0.3471          & 0.9187          & 0.9096          & 11.25          & 183.2          & 6.209          & 0.9945          \\
			& i-ResNet      & 0.0437          & 0.5789          & 0.9205          & 0.9122          & 1.635          & 18.375         & 9.875          & \textbf{0.9974} \\
			& i-ML-Enc(L-1) & \textbf{0.0253} & \textbf{0.3058} & \textbf{0.9934} & \textbf{0.9927} & \textbf{1.165} & \textbf{4.974} & \textbf{5.461} & 0.9876          \\
			\hline
			\multirow{10}{*}{FMNIST}
			& t-SNE          & -               & -               & 0.9896          & 0.9863          & 3.247          & 108.3          & 48.07          & 0.7249          \\
			& ML-Enc         & -               & -               & 0.9903          & 0.9896          & 1.358          & 89.65          & 25.18          & 0.7629          \\
			& AE             & 0.2078          & 27.45           & 0.9744          & 0.9689          & 6.728          & 102.1          & 21.98          & 0.7495          \\
			& TopoAE         & 0.2236          & 31.01           & 0.9658          & 0.9813          & 6.982          & 115.4          & 23.53          & 0.7503          \\
			& ML-AE          & 0.4912          & 18.84           & 0.9912          & \textbf{0.9917} & 1.738          & 101.7          & 25.89          & \textbf{0.7665} \\
			& i-ML-Enc (L)   & \textbf{0.0461} & \textbf{0.3567} & \textbf{0.9923} & 0.9905          & \textbf{1.295} & \textbf{83.63} & \textbf{20.13} & 0.7644          \\ \cline{2-10}
			& INN            & 0.0627          & 0.6819          & 0.9832          & 0.9744          & 1.364          & 21.36          & 9.258          & 0.8471          \\
			& i-RevNet       & 0.0475          & \textbf{0.3519} & 0.9157          & 0.8967          & 21.58          & 204.8          & 6.517          & 0.9386          \\
			& i-ResNet       & 0.0582          & 0.6719          & 0.9242          & 0.9058          & 1.953          & 22.75          & 9.687          & \textbf{0.9477} \\
			& i-ML-Enc(L-1) & \textbf{0.0461} & 0.3567          & \textbf{0.9935} & \textbf{0.9959} & \textbf{1.356} & \textbf{6.704} & \textbf{6.017} & 0.8538           \\
			\hline
			\multirow{10}{*}{Coil-20}
			& t-SNE         & -               & -               & 0.9911          & \textbf{0.9954} & 5.794          & 101.2          & 17.22          & 0.9039          \\
			& ML-Enc        & -               & -               & 0.9920          & 0.9889          & 1.502          & 70.79          & \textbf{9.961} & \textbf{0.9564} \\
			& AE            & 0.3507          & 24.09           & 0.9745          & 0.9413          & 4.524          & 85.09          & 11.45          & 0.8958          \\
			& TopoAE        & 0.4712          & 26.66           & 0.9768          & 0.9625          & 5.272          & 98.33          & 27.19          & 0.9043          \\
			& ML-AE         & 0.1220          & 16.86           & 0.9914          & 0.9885          & \textbf{1.489} & \textbf{68.63} & 10.34          & 0.9548          \\
			& i-ML-Enc (L)  & \textbf{0.0312} & \textbf{1.026}  & \textbf{0.9921} & 0.9871          & 1.695          & 71.86          & 11.13          & 0.9386          \\ \cline{2-10} 
			& INN           & 0.0758          & 0.8075          & 0.9791          & 0.9681          & 2.033          & 79.25          & 8.595          & 0.9936          \\
			& i-RevNet      & 0.0508          & 0.7544          & 0.9316          & 0.9278          & 11.34          & 147.2          & 9.803          & \textbf{1.000}  \\
			& i-ResNet      & 0.0544          & \textbf{0.7391} & 0.9258          & 0.9136          & 1.821          & 13.56          & 10.41          & \textbf{1.000}  \\
			& i-ML-Enc(L-1) & \textbf{0.0312} & 0.9263          & \textbf{0.9940} & \textbf{0.9937} & \textbf{1.297} & \textbf{4.439} & \textbf{7.539} & \textbf{1.000}  \\
			\bottomrule
		\end{tabular}}
	\end{center}
\end{table}

\textbf{Results on toy datasets.} \quad
The Tab. \ref{table:A_2_toy} compares the \textit{i-ML-Enc} with other methods in 9 performance metrics on Swiss roll and Half-spheres datasets in the case of $s=s'=d$. Eight methods for manifold learning: Isomap \cite{2000-science-Isomap}, t-SNE \cite{2008-JMLR-tSNE}, RR \cite{2016-NIPS-RR}, and ML-Enc \cite{2020-MLDL} are compared for NLDR; four AE-based methods AE \cite{2006-science-AE}, VAE \cite{2014-ICLR-VAE}, GRAE \cite{2020-GRAE}, TopoAE \cite{2020-ICML-TopoAE}, and ML-AE \cite{2020-MLDL} are compared for reconstructible manifold learning. We report the $L$-th layer of \textit{i-ML-Enc} (the first stage) for the NLDR quality and the $(L-1)$-th layer (the second stage) for the invertible NLDR ability. ML-Enc performs best in Trust, $K$min, $K$max, and $l$-MSE on Swiss roll which shows its great embedding abilities. Based on ML-Enc, \textit{i-ML-Enc} achieves great embedding results in the second stage on Half-spheres, which shows its advantages of preserving topological and geometric structures in the high-dimensional case. And \textit{i-ML-Enc} outperforms other methods in its invertible NLDR property of the first stage.

\begin{figure}[!htb]
  \centering
  \includegraphics[width=4.8in]{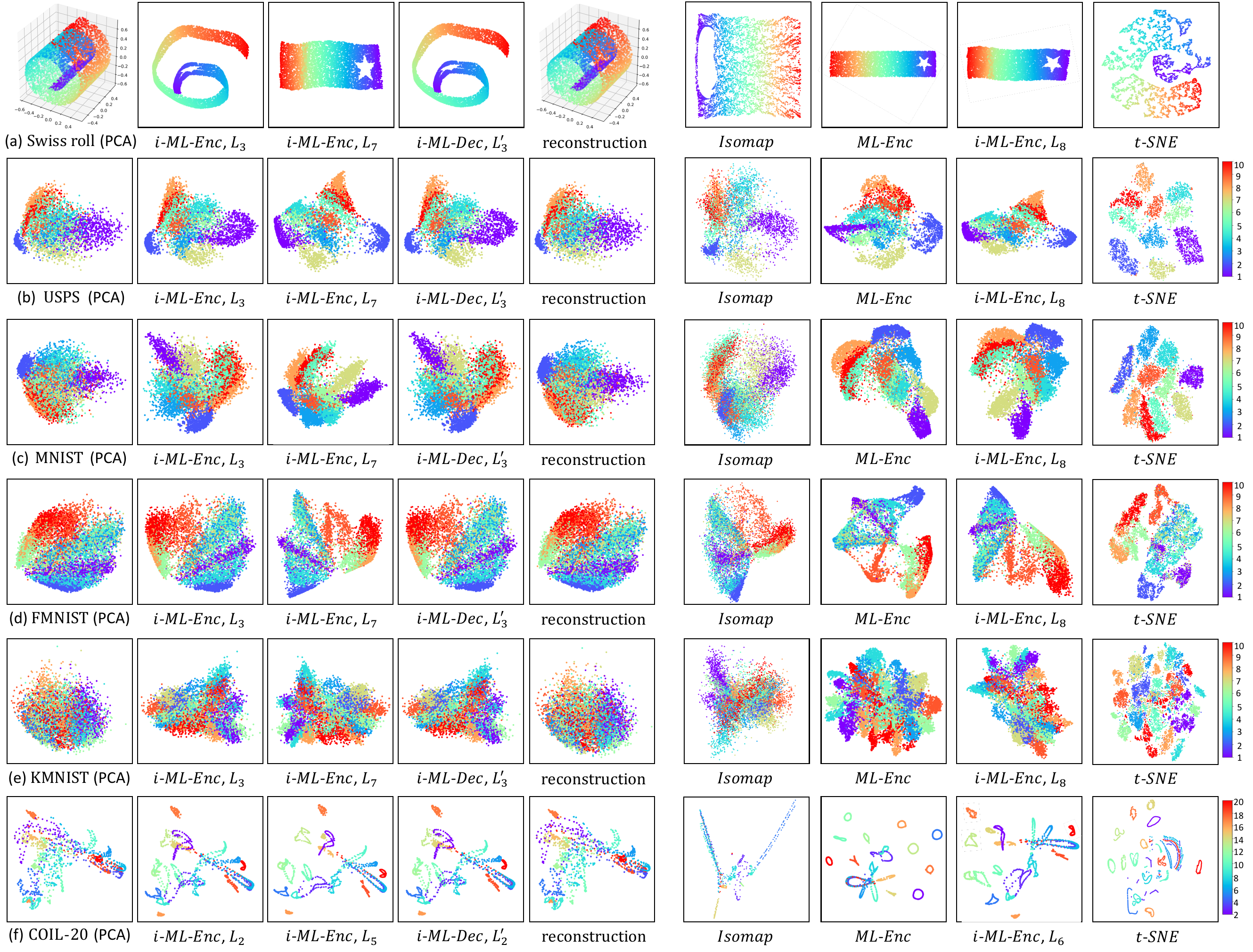}
  \caption{Visualization of invertible NLDR results of \textit{i-ML-Enc} with comparison to Isomap, ML-Enc, and t-SNE on Swiss roll and five real-world datasets. The target dimension $s'=2$ for all datasets, and the high-dimensional latent space are visualized by PCA. For each row, the left five cells show the NLDR and reconstruction process in the first stage of \textit{i-ML-Enc}, and the right four cells show 2D results for comparison. ML-Enc and t-SNE show great clustering effects but drop topological information. Compared to the classical DR method Isomap (preserving the global geodesic distance) and t-SNE (preserving the local geometry), the representations learned by \textit{i-ML-Enc} preserves the relationship between clusters and the local geometry within clusters.}
  \label{figure:A2_real_full}
\end{figure}

\begin{figure}[!htb]
  \centering
  \includegraphics[width=4.6in]{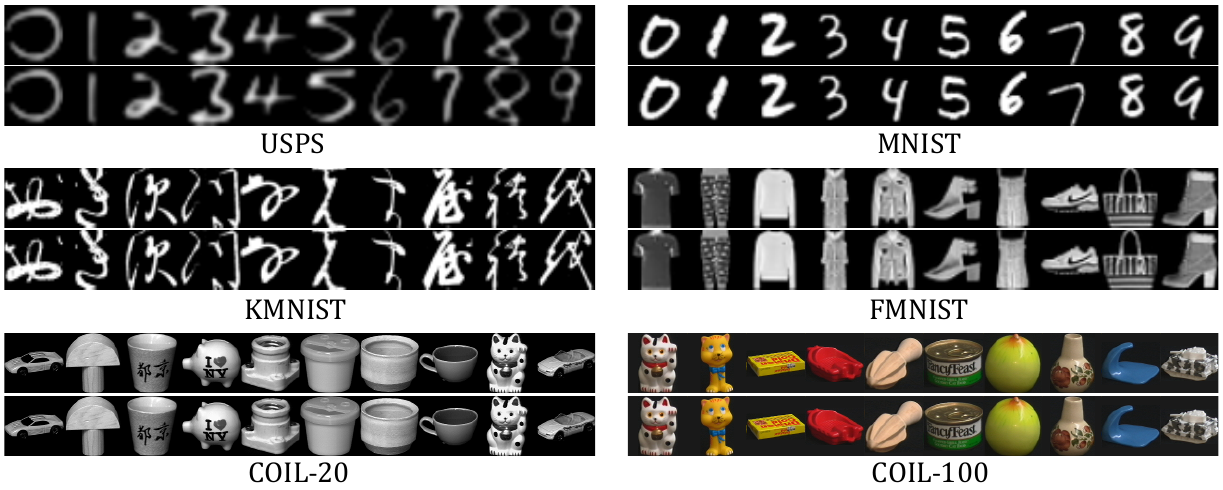}
  \caption{Reconstruction results of \textit{i-ML-Enc} on six image datasets. For each cell, the upper row shows results of \textit{i-ML-Enc} while the lower row shows the raw input images. We randomly selected 10 images from different classes to show the bijective property of \textit{i-ML-Enc}.}
  \label{figure:A2_reconstruction}
\end{figure}

\textbf{Results on real-world datasets.} \quad
The Tab. \ref{table:A_2_real} compares the \textit{i-ML-Enc} with other methods in 9 performance metrics on USPS, FMNIST and COIL-20 datasets in the case of $s>s'$. Six methods for manifold learning: Isomap, t-SNE, and ML-Enc are compared for NLDR; three AE-based methods AE, ML-AE, and TopoAE are compared for reconstructible manifold learning. Three methods for inverse models: INN \cite{2019-GCPR-auto_INN}, i-RevNet \cite{2018-ICLR-iRevNet}, and i-ResNet \cite{2019-ICML-iResNet} are compared for bijective property. The visualization of NLDR and its inverse process of \textit{i-ML-Enc} are shown in Fig. \ref{figure:A2_real_full}, together with the NLDR results of Isomap, t-SNE and, ML-Enc. The target dimension for visualization is $s^{'}=2$, and the high-dimensional latent space is visualized by PCA. Compared with NLDR algorithms, the representation of the $L$-th layer of \textit{i-ML-Enc} nearly achieves the best NLDR metrics on FMNIST, and ranks second place on USPS and third place on COIL-20. The drop of performance between the $(L-1)$-th and $L$-th layers of \textit{i-ML-Enc} are caused by the sub-optimal linear transformation layer, since the representation of its first stage is quite reliable. Compared with other inverse models, \textit{i-ML-Enc} outperforms in all the NLDR metrics and inverse metrics in the first stage, which indicates that a great low-dimensional representation of data manifolds can be learned by a series of equidimensional layers. However, \textit{i-ML-Enc} shows larger NME on FMNIST and COIL-20 compared with inverse models, which indicates that \textit{i-ML-Enc} is more unstable dealing with complex datasets in the first stage. Besides, the reconstruction samples from six image datasets including COIL-100 \cite{1996-coil100} show the inverse quality of \textit{i-ML-Enc} in Fig. \ref{figure:A2_reconstruction}.


\subsection{Latent Space Interpolation}
\subsubsection{Datasets Comparison}
\label{A_3.1}
Here is a brief introduction to four interpolation data sets, as shown in Tab. \ref{table:A3_interp_dataset_intro}. We analyze the difficulty of dataset roughly according to \textbf{dimension}, \textbf{sample size}, \textbf{image entropy}, \textbf{texture}, and the performance of \textbf{classification tasks}: (1) Sampling ratio. Generally, the sample number has an exponential relationship with the input dimension in the case of sufficient sampling. Thus, the sampling ratio of USPS is higher than others. (2) Image entropy. The Shannon entropy of the histogram measures the information content of the image, and it reaches the maximum when the density estimated by the histogram is a uniform distribution. We report the mean entropy of each dataset. We conclude that USPS has richer grayscale than MNIST(256), while the information content of MNIST(784), KMNIST, and FMNIST shows an increasing trend. (3) Texture. The standard deviation (std) of the histogram reflects the texture information in the image, and we report the mean std of each dataset. Combined with the evaluation of human eyes, the texture features become rougher and rougher from USPS, MNIST, to KMNIST, while FMNIST contains complex and regular textures. (4) Classification tasks. We report the mean accuracy of $10$-fold cross-validation using kNN and logistic classifier \cite{2011-JMLR-sklearn} for each data set. The credibility of the neighborhood system decreases gradually from USPS, MNIST, KMNIST to FMNIST. Combined with the visualization results in Fig. \ref{figure:A2_real_full}, it obvious that KMNIST has the worst linear separability. 
Above all, we roughly order the difficulty of manifold learning on each data set: \textbf{USPS$<$MNIST(256)$<$MNIST(784)$<$KMNIST$<$FMNIST}.

\begin{table}[htb]
	\setlength{ \abovecaptionskip}{0.cm}
	\caption{Comparison of manifold learning difficulties of interpolation datasets. For each dataset, we report entropy and std (Texture) on the entire image histogram, and mean accuracy of $10$-fold cross-validation of the kNN classifier.}
	\label{table:A3_interp_dataset_intro}
	\begin{center}
	\resizebox{0.925\columnwidth}{!}{
		\begin{tabular}{lllllll}
			\toprule
			Dataset    & Sample    & Dimension & Entropy & Texture & KNN    \\
			\hline
			USPS       & 9298      & 256       & 5.479   & 0.5097  & 0.9589 \\
			MNIST(256) & 9298      & 256       & 1.879   & 10.51   & 0.9493 \\
			MNIST(784) & 20000     & 784       & 1.598   & 39.75   & 0.9515 \\
			KMNIST     & 20000     & 784       & 2.911   & 33.01   & 0.9141 \\
			FMNIST     & 20000     & 784       & 4.115   & 24.75   & 0.8133 \\
			\bottomrule
		\end{tabular}}
	\end{center}
\end{table}
\begin{figure}[!htb]
  \centering
  \includegraphics[width=4.8in]{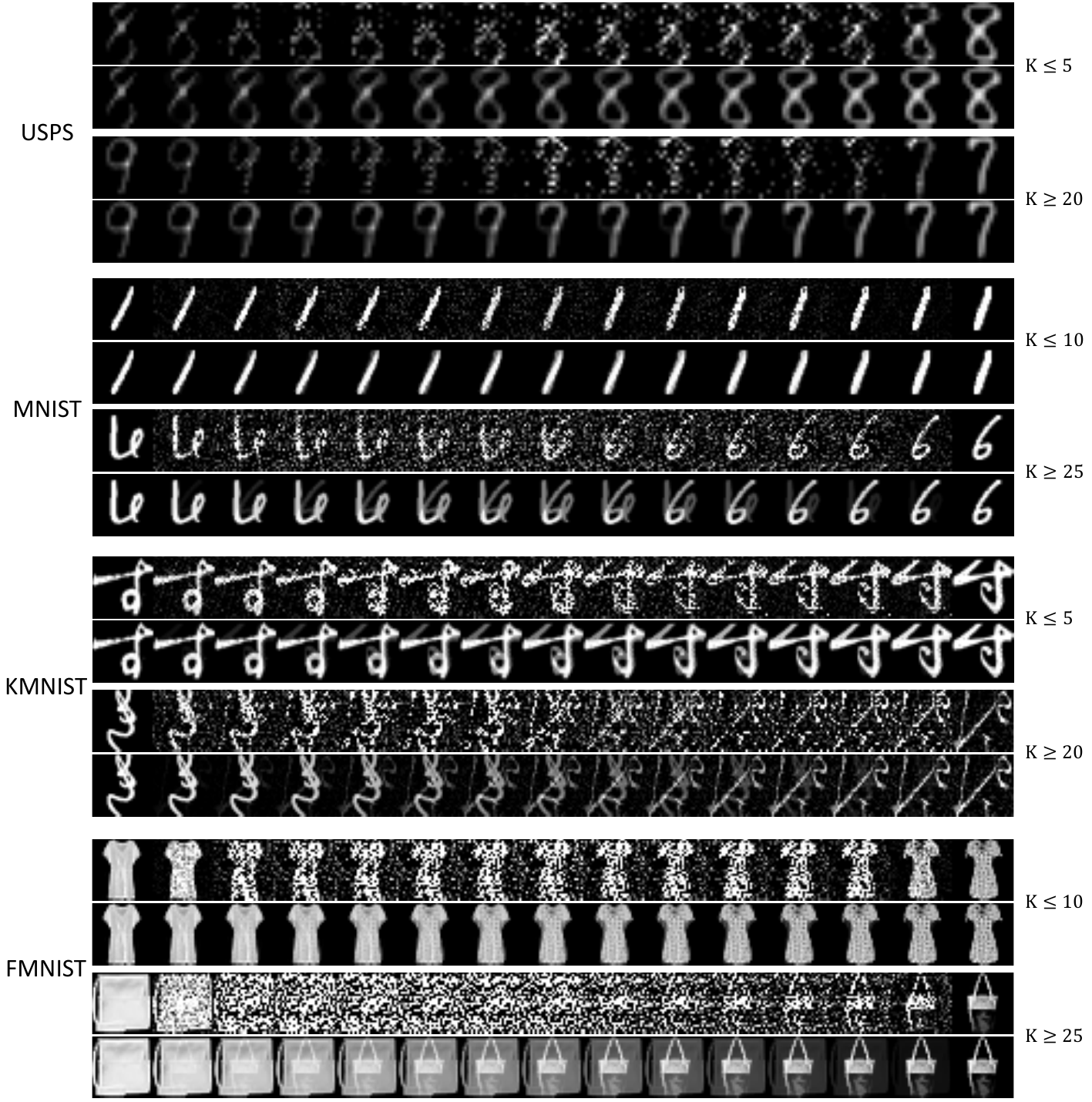}
  \caption{Visualization of kNN interpolation results of \textit{i-ML-Enc} on image datasets with $k\le10$ and $k\ge20$. For each row, the upper part shows results of \textit{i-ML-Enc} while the lower part shows the raw input images. Both the input and latent results transform smoothly when $k$ is small, while the latent results show more noise but less overlapping and pseudo-contour than the input results when $k$ is large. The latent interpolation results show more noise and less smoothness when the data manifold becomes more complex.}
  \label{figure:A3_inter_kNN}
\end{figure}
\begin{figure}[!htb]
	\centering
	\includegraphics[width=4.6in]{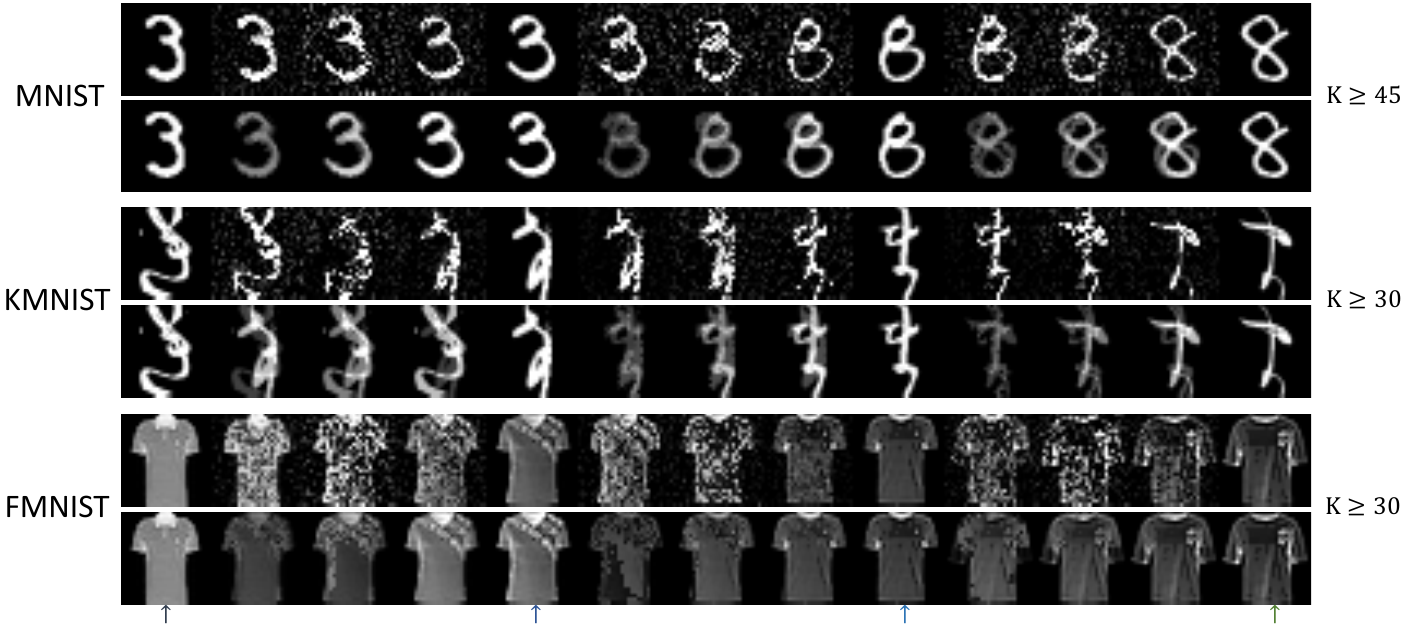}
	\caption{The interpolation results of the geodesic interpolation in the latent space. For each dataset, the upper row shows the latent result, while the lower shows the input result. The samples 1, 5, 9, 13 pointed by the arrow are the original samples.}
	\label{figure:A3_inter_jump}
\end{figure}

\subsubsection{More Interpolation Results}
\label{A_3.2}
\textbf{kNN interpolation.} \quad
We verify the reliability of the low-dimensional representation by kNN interpolation. Comparing the results of different values of $k$, as shown in Fig. \ref{figure:A3_inter_kNN}, we conclude that: 
(1) Because the high-dimensional latent space is still quite sparse, there is some noise caused by linear approximation on the latent results. The MSE loss and noises of the latent results are increasing with the expansion of the local neighborhood on the same dataset, reflecting the reliability of the local neighborhood system. 
(2) In terms of the same sampling rate, the MSE loss and noises of the latent results grow from MNIST(784), KMNIST to FMNIST, which indicates that the confidence of the local homeomorphism property of the latent space decreases gradually on more difficult manifolds. 
(3) In terms of the similar data manifolds, USPS(256) and MNIST(256) show better latent interpolation results than MNIST(784), which demonstrates that it is harder to preserve the geometric properties on higher input dimension. 
(4) Though the latent results import some noise, the input results have unnatural transformations such as pseudo-contour and overlapping. Thus, the latent space results are more smooth than the input space, which validates that the latent space learned by \textit{i-ML-Enc} is flatter and denser than the input space. In a nutshell, we infer that the difficulty of preserving the geometric properties based on approximation of the local tangent space by the local neighborhood is the crucial reason for the manifold learning failure in the real-world case.

\textbf{Geodesic interpolation.} \quad
We further perform the latent interpolation along the geodesic path between sample pairs when $k$ is large to generate reliable intermediate samples. It might reflect the topological structure of data manifolds when two samples in a sample pair are in different clusters. Given a sample pair $(x_{i},x_{j})$ with $k\ge45$ from different clusters, we select the three intermediate sample pairs $(x_{i},x_{i_1})$, $(x_{i_1},x_{i_2})$, $(x_{i_2},x_{j})$ with $k\le20$ along the geodesic path in latent space for piece-wise linear interpolation in both space. Compared with results of MNIST, KMNIST, and FMNIST, as shown in Fig. \ref{figure:A3_inter_jump}, we can conclude: 
(1) The latent results are more reliable than those in the input space which can generate the synthetic samples between two different clusters. 
(2) Compared with MNIST, KMNIST, and FMNIST, the latent results of more complex datasets are more ambiguous and noisy, which indicates that it is more difficult to find a low-dimensional representation of more complex data manifolds with all geometric structures preserved.

\subsection{Analysis of the loss terms}
\label{A_4}
We further conduct ablation study of the \textit{extra head} (\textbf{+Ex}), the orthogonal loss $\mathcal{L}_{orth}$ (\textbf{+Orth}), and the zero padding loss $\mathcal{L}_{pad}$ (\textbf{+Pad}) on MNIST, USPS, KMNIST, FMNIST and COIL-20. The Tab. \ref{table:A_4_ablation} reports ablation results in the 8 indicators and the $r(Z^{L-1})$. We analyze and conclude:  
(1) The combination of \textbf{Ex} and \textbf{Orth} nearly achieve the best inverse and DR performance on MNIST, USPS, FMNIST, and COIL-20, which indicates that it is the basic factor for invertible NLDR in the first $L-1$ layers. (2) When only use \textbf{Orth}, the NLDR in the first $L-1$ layer of the network will degenerate into the identity mapping, and DR is achieved with the linear project on layer $L$. (3) Combined with all three items \textbf{Ex}, \textbf{Orth} and \textbf{Pad}, \textit{i-ML-Enc} obtains a sparse coordinate representation, but achieves little worse embedding quality on USPS and COIL-20 than using \textbf{Ex} and \textbf{Orth}. (4) Besides the proposed loss items, ML-AE overperforms the other combinations in the \textbf{Acc} metric indicating the reconstruction loss helps improve the generation ability of ML-Enc. Above all, the \textbf{Ex+Orth+Pad} combination, i.e. \textit{i-ML-Enc}, can achieve the proposed invertible NLDR.

\renewcommand\tabcolsep{5.0pt}
\begin{table}[!htb]
    \caption{Ablation study of the proposed loss terms in \textit{i-ML-Enc} on five image datasets.}
	\label{table:A_4_ablation}
	\begin{center}
	\resizebox{0.98\columnwidth}{!}{
	\begin{tabular}{c|llllllllll}
		\toprule
		\multicolumn{1}{l}{Dataset}
		& Algorithm    & RMSE            & MNE             & Trust           & Cont            & $K$min         & $K$max         & Acc             & $r(Z^{L-1})$ \\
		\hline
		\multirow{6}{*}{MNIST}
		& ML-AE        & 0.4012          & 16.84           & 0.9893          & 0.9926          & 1.704  	    & 57.48    		 & \textbf{0.9340} & 15           \\
		& ML-Enc       & -               & -               & 0.9862          & \textbf{0.9927} & 1.761          & 58.91          & 0.9326          & 14           \\
		& +Ex          & -               & -               & 0.9891          & 0.9812          & 2.745          & 78.88          & 0.9316          & \textbf{12}  \\
		& +Ex+Orth     & 0.0341          & 0.4255          & 0.9874          & \textbf{0.9927} & 1.817          & 59.97  	     & 0.9298          & 361          \\
		& +Ex+Orth+Pad & 0.0457          & 0.5085          & \textbf{0.9906} & 0.9912          & 2.033          & 60.14          & 0.9316          & 125          \\
		& +Orth        & \textbf{0.0056} & \textbf{0.1275} & 0.9652          & 0.9578          & \textbf{1.597} & \textbf{53.21} & 0.8807          & 716          \\
		\hline
		\multirow{6}{*}{USPS}
		& ML-AE        & 0.4912          & 11.84           & 0.9879          & \textbf{0.9905} & 1.529          & 55.32          & \textbf{0.9576} & 16           \\
		& ML-Enc       & -               & -               & 0.9874          & 0.9897          & 1.562          & \textbf{52.14} & 0.9534          & 14           \\
		& +Ex          & -               & -               & 0.9849          & 0.9836          & 2.525          & 78.88          & 0.9413          & \textbf{11}  \\
		& +Ex+Orth     & 0.0395          & 0.2511          & \textbf{0.9895} & 0.9875          & 1.366          & 58.83          & 0.9376          & 192          \\
		& +Ex+Orth+Pad & 0.0253          & 0.3058          & 0.9886          & 0.9861          & 1.538          & 60.79          & 0.9456          & 116          \\
		& +Orth        & \textbf{0.0109} & \textbf{0.2043} & 0.9702          & 0.9654          & \textbf{1.328} & 66.25          & 0.8961          & 243          \\
		\hline
		\multirow{6}{*}{KMNIST}
		& ML-AE        & 0.4912          & 18.84           & 0.9781          & \textbf{0.9912} & 2.478          & 80.66          & 0.7639          & 19           \\
		& ML-Enc       & -               & -               & 0.9738          & 0.9883          & 2.253          & 103.4          & \textbf{0.7719} & \textbf{18}  \\
		& +Ex          & -               & -               & 0.9786          & 0.9801          & 5.826          & 255.1          & 0.7624          & \textbf{18}  \\
		& +Ex+Orth     & 0.0463          & 0.4661          & 0.9805          & 0.9872          & 2.396          & 70.89          & 0.6325          & 406          \\
		& +Ex+Orth+Pad & 0.0844          & 0.4589          & \textbf{0.9875} & 0.9894          & 2.697          & 78.19          & 0.7513          & 198          \\
		& +Orth        & \textbf{0.0223} & \textbf{0.1962} & 0.9621          & 0.9593          & \textbf{1.991} & \textbf{60.51} & 0.5922          & 732          \\
		\hline
		\multirow{6}{*}{FMNIST}
		& ML-AE        & 0.4912          & 18.84           & 0.9912          & \textbf{0.9917} & 1.738          & 101.7          & \textbf{0.7665} & 19           \\
		& ML-Enc       & -               & -               & 0.9903          & 0.9896          & 1.358          & 89.65          & 0.7629          & 18           \\
		& +Ex          & -               & -               & 0.9886          & 0.9726          & 5.826          & 279.4          & 0.7624          & \textbf{16}  \\
		& +Ex+Orth     & 0.0337          & 0.3194          & 0.9895          & 0.9840          & 1.879          & 98.66          & 0.7613          & 393          \\
		& +Ex+Orth+Pad & 0.0461          & 0.3567          & \textbf{0.9923} & 0.9905          & \textbf{1.298} & \textbf{83.63} & 0.7644          & 182          \\
		& +Orth        & \textbf{0.0152} & \textbf{0.2975} & 0.9701          & 0.9593          & 2.073          & 89.03          & 0.5934          & 743          \\
		\hline
		\multirow{4}{*}{COIL-20}
		& ML-AE        & 0.1220          & 16.87           & 0.9914          & 0.9885          & 1.489          & 74.79          & \textbf{0.9564} & \textbf{44}  \\
		& ML-Enc       & -               & -               & 0.9920          & \textbf{0.9889} & 1.502          & 70.79          & \textbf{0.9564} & 46           \\
		& +Ex+Orth     & \textbf{0.0049} & \textbf{0.093}  & \textbf{0.9927} & 0.9852          & \textbf{1.378} & \textbf{66.39} & 0.9427          & 1190         \\
		& +Ex+Orth+Pad & 0.0171          & 1.026           & 0.9921          & 0.9871          & 1.695          & 71.86          & 0.9386          & 746          \\
		\bottomrule
	\end{tabular}}
	\end{center}
\end{table}





\end{document}